\documentclass[11pt]{article}

\usepackage[top=1in,bottom=1in, left=1in, right=1in]{geometry} 

\usepackage{amsmath,amsthm,amssymb,amsopn,amsfonts,pdfpages,algorithmic,algorithm,dsfont}
\usepackage{graphicx}
\usepackage{authblk}
\usepackage{blkarray}
\DeclareGraphicsExtensions{.pdf,.jpeg,.png}
  \usepackage[caption=false,font=footnotesize,labelfont=sf,textfont=sf]{subfig}
  \usepackage[colorlinks=true,linkcolor=cyan,citecolor=blue]{hyperref}
\usepackage{blkarray}
\usepackage{multirow}

\DeclareMathOperator*{\argmin}{arg\,min}
\DeclareMathOperator*{\argmax}{arg\,max}

\newcommand{\fl}{\left\lfloor \frac{n/3}{\xi_n} \right\rfloor}
\newcommand{\va}{V_{\mathcal{A}}}
\newcommand{\fa}{f_{\mathcal{A}}}

\newcommand{\VNA}{\text{VN}\circ\text{GMM}\circ\text{ASE}}
\newcommand{\nn}{\mathfrak{N}_{\bf V^*}}
\newcommand{\rp}{\binom{\text{precision}}{\text{recall}}}

\newcommand{\gn}{\mathcal{G}_n}

\newcommand{\gm}{\mathcal{G}_m}

\newcommand{\gnma}{\mathcal{G}_{n,m}^a}

\newcommand{\calT}{{\mathcal T}}

\newcommand{\p}{\mathbb{P}}
\newcommand{\e}{\mathbb{E}}

\newcommand{\mO}{\mathfrak{O}}

\newcommand{\rank}{\text{rank}}
\newcommand{\PF}{\p_{ F_{c,\theta}^{(n,m)} } }

\newcommand{\fo}{\mathfrak{o}}

\newcommand{\mo}{\mathfrak{o}}

\newcommand{\vv}{\textcolor{black}}
\newtheorem{theorem}{Theorem}

\newtheorem{lemma}[theorem]{Lemma}

\theoremstyle{definition}
\newtheorem{definition}[theorem]{Definition}
\theoremstyle{remark}
\newtheorem{remark}[theorem]{Remark}

\begin{document}
\title{Vertex Nomination, Consistent Estimation, and
Adversarial Modification}
\author[$1$]{Joshua Agterberg}
\author[$2$]{Youngser Park}
\author[$3$]{Jonathan Larson}
\author[$3$]{Christopher White}
\author[$1,2$]{Carey E. Priebe}
\author[$4$]{Vince~Lyzinski}

\affil[$1$]{\small Department of Applied Mathematics and Statistics, Johns Hopkins University}
\affil[$2$]{\small Center for Imaging Sciences, Johns Hopkins University}
\affil[$3$]{\small Microsoft AI and Research,
Microsoft}
\affil[$4$]{\small Department of Mathematics, University of Maryland}

\maketitle
\begin{abstract}
Given a pair of graphs $G_1$ and $G_2$ and a vertex set of interest in $G_1$, the vertex nomination (VN) problem seeks to find the corresponding vertices of interest in $G_2$ (if they exist) and produce a rank list of the vertices in $G_2$, with the corresponding vertices of interest in $G_2$ concentrating, ideally, at the top of the rank list.  In this paper, we define and derive the analogue of Bayes optimality for VN with multiple vertices of interest, and we define the notion of maximal consistency classes in vertex nomination. This theory forms the foundation for a novel VN adversarial contamination model, and we demonstrate with real and simulated data that there are VN schemes that perform effectively in the uncontaminated setting, and  adversarial network contamination adversely impacts the performance of our VN scheme.
We further define a network regularization method for mitigating the impact of the adversarial contamination, and we demonstrate the effectiveness of regularization in both real and synthetic data.
\end{abstract}


\section{Introduction and Background}\label{sec:intro}
Given graphs $G_1$ and $G_2$ and vertices of interest $V^*\subset V(G_1)$, the aim of the vertex nomination (VN) problem is to rank the vertices of $G_2$ into a nomination list with the corresponding vertices of interest concentrating at the top of the nomination list.
In recent years, a host of VN procedures have been introduced (see, for example, \cite{coppersmith2012vertex,marchette2011vertex,LeePri2012,FisLyzPaoChePri2015,patsolic2017vertex,yoder2018vertex}) that have proven to be effective information retrieval tools in both synthetic and real data applications.
Moreover, recent work establishing a fundamental statistical framework for VN has led to a novel understanding of the limitations of VN efficacy in evolving network environments \cite{lyzinski2017consistent}.  
Herein, we consider a general statistical model for adversarial contamination in the context of vertex nomination---here the adversary model can both randomly add or remove edges and/or vertices in the network ---and we examine the effect of both these contaminations on VN performance.  
In addition, we extend existing theory on consistent vertex nomination to multiple vertices of interest and define and derive Bayes Optimal Classifiers in this setting.  We further show that there are infinitely many classes of distribution for which a vertex nomination scheme is not consistent.  

\textcolor{black}{The practical additional value of this paper is to \begin{enumerate}
    \item extend the results of \cite{lyzinski2017consistent} to the more realistic multiple VOI setting;
    \item rigorously frame the concept of an adversary in the random graph framework;
    \item develop theory showing how it is possible for an adversary to render vertex nomination schemes inconsistent;
    \item demonstrate empirically that although an adversary can have a negative impact, regularization can succeed in recovering consistency.  
\end{enumerate}
The reason we do not prove that regularization succeeds is that the regularization scheme depends on the particular graph observation and introduces complex dependence structure into the problem.  Such dependence, coupled with the already difficult spectral analysis problem, makes it unclear what exactly is even being estimated when using any spectral nomination scheme with regularization.  Furthermore, the regularization scheme we consider is highly model-dependent, and our main theoretical contributions apply to \emph{any} vertex nomination scheme and as such are necessary to begin to understand adversarial vertex nomination. 
}

To motivate our mathematical and statistical results further, we first consider an illustrative real data example in Section \ref{sec:Motive} in which we demonstrate the following: A VN scheme that works effectively with network contamination adversely impacting the performance of our VN scheme. 
Note that we will provide a more thorough background of the relevant literature after the motivating example in Section \ref{sec:BG}.

\subsection{Motivating example}
\label{sec:Motive}

Consider the pair of high school friendship networks in \cite{mastrandrea2015contact}: The first, $G_1$, has $156$ nodes, each representing a student, and has two vertices adjacent if the two students made contact with each other at school in a given time period; the second, $G_2$, has $134$ vertices, again with each vertex representing a student, and has two vertices adjacent if the two students are friends on Facebook.  
There are $82$ students appearing in both $G_1$ and $G_2$, and we pose the VN problem here as follows: given a student-of-interest in $G_1$, can we nominate the corresponding student (if they exist) in $G_2$.
We note here that the vertex nomination approach outlined below easily adapts to the multiple vertices of interest (v.o.i.) scenario (i.e., given students-of-interest in $G_1$, can we nominate the corresponding students, if they exist, in $G_2$)---and we will provide the necessary details for handling both single and multiple v.o.i.\@ below.  Recall that the VN problem assumes there is a correspondence between the vertices but that the practitioner does not have access to this correspondence.  To this end, we act as though we do not know the corresponding student in each graph.

In one idealized data setting, all students would appear in both graphs as this would potentially maximize the signal present in the correspondence of labels across graphs.
This bears itself out in the following illustrative VN experiment.
Consider the following simple VN scheme, which we denote $\VNA$:
Given vertex (or vertices) of interest $v^*$ in $G_1$ and seeded vertices $S\subset V_1\cap V_2$ (seeds here represent vertices whose identity across networks is known a priori), we proceed by embedding the graphs into a common Euclidean space $\mathbb{R}^d$ and clustering using Mahalanobis distances between the embeddings of the vertices (see Section \ref{sec:asegmm} for full detail).%

\begin{figure*}[!t]
\centering
\subfloat{\includegraphics[width=0.45\textwidth]{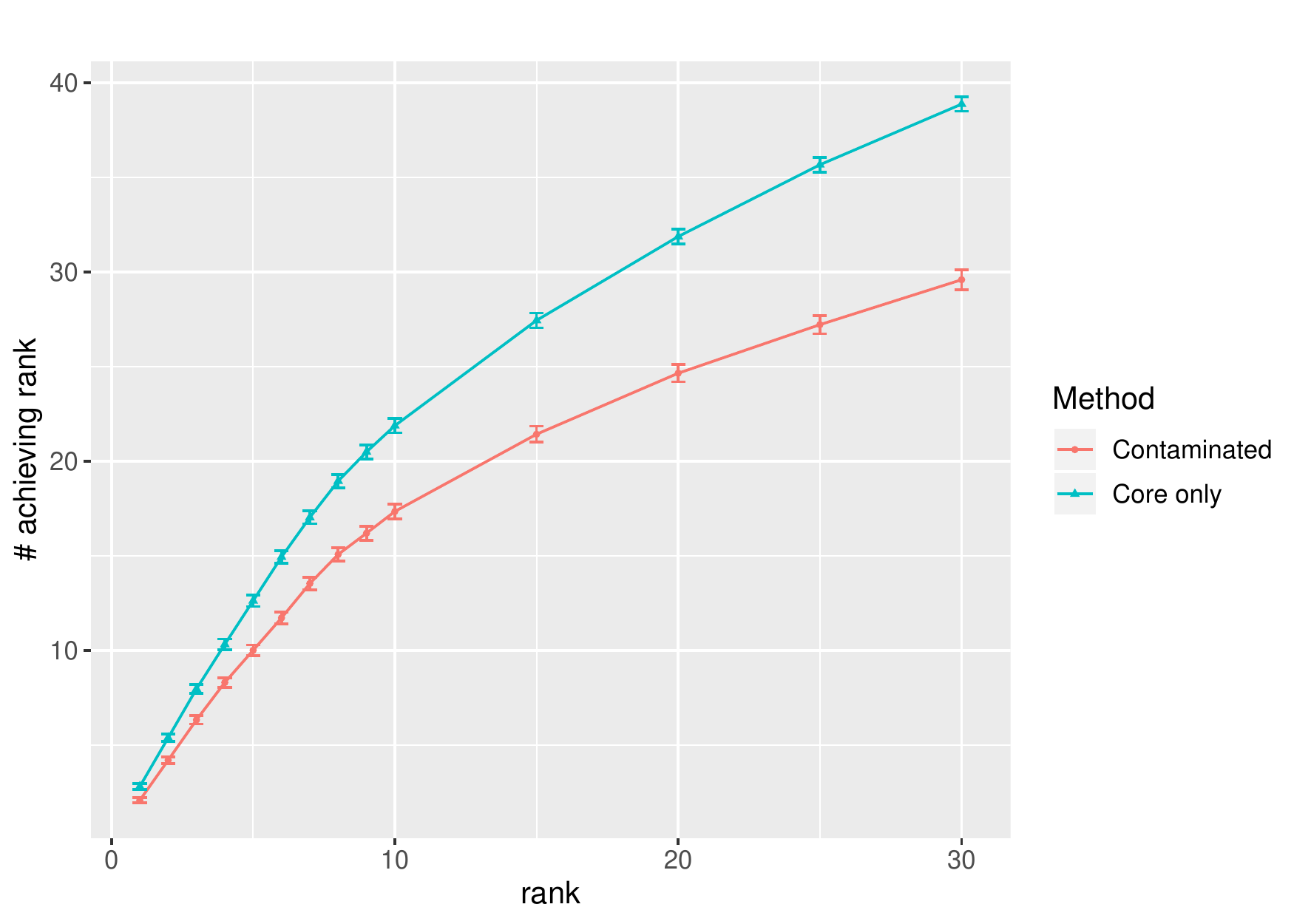}
\label{fig_first_case}}
\hfil
\subfloat{\includegraphics[width=0.45\textwidth]{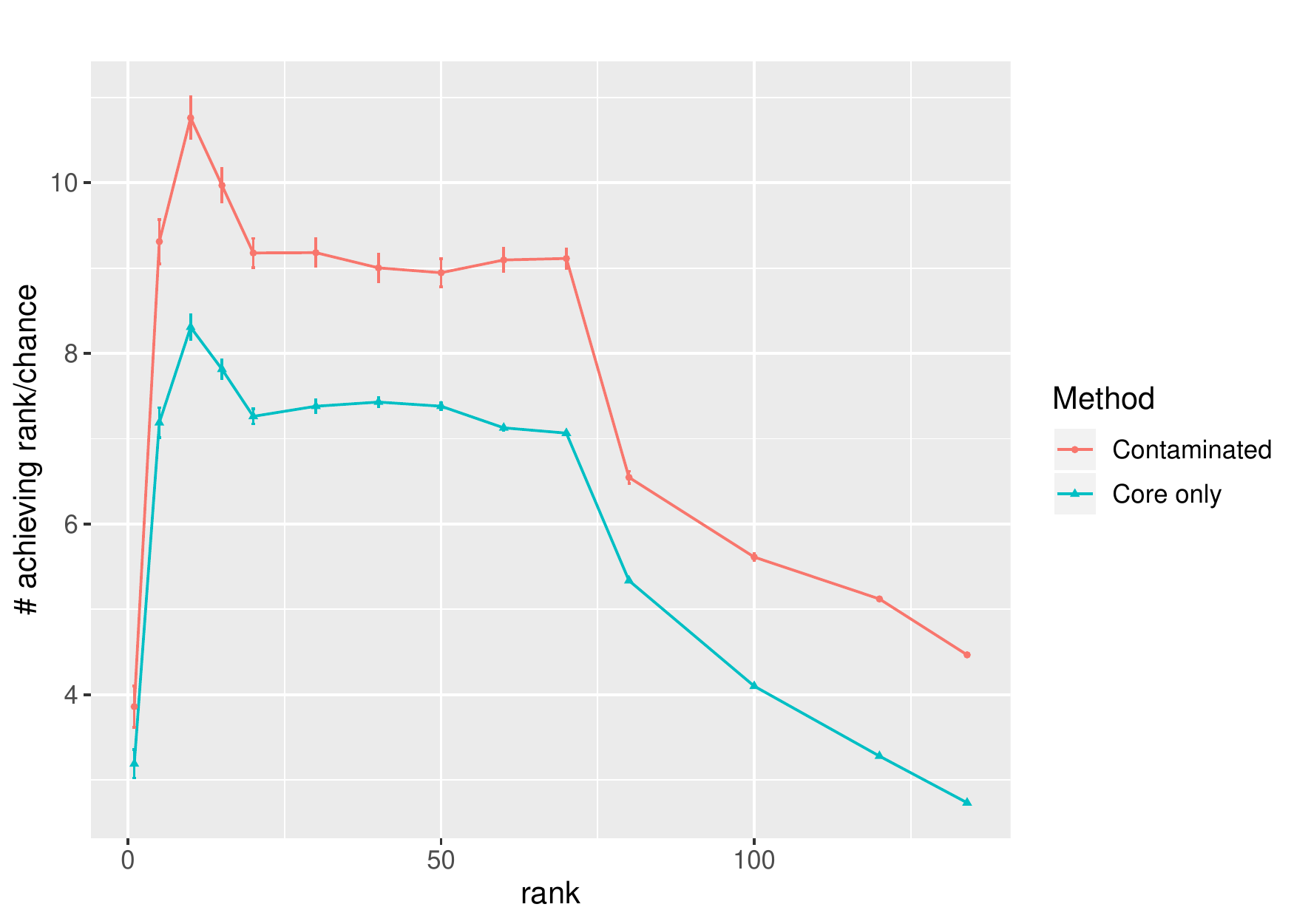}
\label{fig_second_case}}
\caption{ 
We plot the performance of $\VNA$ averaged over $nMC=500$ random seed sets of size $s=10$.   The left figure shows the number of true vertices achieving the rank, and the right figure shows the same result normalized by chance performance.  
The blue line represents performance with the truth, and the red line represents the contaminated network. 
See Section \ref{sec:Motive} for more details.}
\label{fig:vnfex}
\end{figure*}

We can consider running the $\VNA$ in the idealized data setting where we only consider the induced subgraphs of $G_1$ and $G_2$ containing the $82$ common vertices across graphs (call these graphs $G_1^{(i)}$ and $G_2^{(i)}$), and we can also consider running the procedure in the setting where the $52$ vertices in $G_2$ without matches across graphs are added to $G_2^{(i)}$ as a form of contamination.
These unmatchable vertices can have the effect of obfuscating the correspondence amongst the common vertices across graphs, and thus can diminish VN performance.
Indeed, we see this play out in Figure \ref{fig:vnfex}.

In Figure \ref{fig:vnfex}, we plot the performance of $\VNA$ averaged over $nMC=500$ random seed sets of size $s=10$.
In the left figure, the $x$-axis shows the ranks in the nomination list and the $y$-axis shows the mean ($\pm$ 2s.e.) number of vertices $v\in G_1^{(i)}$, when viewed as the lone v.o.i., that had their corresponding vertex of interest ranked in the top $x$ by $\VNA$. 
The right figure shows the same results normalized by chance performance, where we plot 
$$y=\tfrac{
\text{mean }\#\text{ of v.o.i. with corresp. v.o.i. ranked in top }x\text{ by }\VNA}
{\text{mean }\#\text{ of v.o.i. with corresp. v.o.i. ranked in top }x\text{ by chance algorithm}
}$$ versus $x$.
The blue line represents performance in the idealized networks $G_1^{(i)}$ and $G_2^{(i)}$, and the red line represents performance in the contaminated network pair $(G_1^{(i)},G_2)$.
We see that the contamination detrimentally affects the performance of $\VNA$ at all levels, as \emph{for all} $x$, the number of v.o.i. in $G_1^{(i)}$ with their corresponding v.o.i. ranked in the top $x$ in the second graph is larger in $(G_1^{(i)},G_2^{(i)})$ versus in $(G_1^{(i)},G_2)$.
Note that the chance normalization is computed separately under the core and noisy models, and the seeming performance gain relative to chance in the contaminated setting is attributable to the fact that $G_2$ has significantly more vertices than the idealized $G_2^{(i)}$, and chance is therefore significantly worse.
We emphasize here the effect of the contamination on VN performance; indeed, the adversarial contamination greatly (negatively) effects the performance of our vertex nomination scheme, suggesting that perhaps the vertex nomination scheme is not consistent for this class of contaminated distributions.  
\vv{In effect, the adversary is knocking the networks out of the consistency class for $\VNA$; see Section \ref{sec:CC} for detail.}
\vv{While the results of Section \ref{sec:verify} show that we cannot verify (in an unsupervised manner, without the true labels) the extent to which the contamination negatively impacts the performance of VN, in Section \ref{sec:regreg}, we empirically explore the impact of regularization strategies for mitigating this contamination.}

\begin{remark}[The role of seeds]
Figure \ref{fig:vnfex} shows performance of $\VNA$ averaged over $500$ randomly chosen seed sets of size $10$.
While performance, on the whole, increases with proper regularization, the story can vary wildly from seed set to seed set. 
While a full exploration of this is beyond the scope of the present text, this is an active area of our work.
\end{remark}

\subsection{Background} 
\label{sec:BG}
In modern statistics and machine learning, graphs are a common way to take into account the complex relationships between data objects, and graphs have been used in applications across the biological (see, for example, \cite{neu1,neu2,neu3,bio1,bio2,bio3}) and social sciences (see, for example, \cite{socnet1,socnet2,resp1,resp2}).
In addition to more traditional statistical inference tasks such as clustering \cite{rohe2011spectral,qin2013dcsbm,networks08:_v,newman2006modularity}, classification \cite{vogelstein2013graph,chen2016robust,neu3}, and estimation \cite{bickel2013asymptotic,BicChe2009,sussman2014consistent}, there has been significant work in more network-specific inference tasks such as 
graph matching \cite{ConteReview,foggia2014graph,yan2016short}, and vertex nomination \cite{marchette2011vertex,coppersmith2014vertex,FisLyzPaoChePri2015}.  

Recall that the vertex nomination problem can be stated loosely as follows: given graphs $G_1$ and $G_2$ and vertices of interest $V^*\subset V(G_1)$, rank the vertices of $G_2$ into a nomination list with the corresponding vertices of interest concentrating at the top of the nomination list (see Definition \ref{def:VN} for full detail).
While vertex nomination has found applications in a number of different areas, such as social networks in \cite{patsolic2017vertex} and data associated with human trafficking in \cite{FisLyzPaoChePri2015}, 
there are relatively few results establishing the statistical properties of vertex nomination.  In \cite{FisLyzPaoChePri2015}, consistency is developed within the stochastic blockmodel random graph framework, where interesting vertices were defined via community membership.
In \cite{lyzinski2017consistent}, the authors develop the concepts of consistency and Bayes optimality  for a very general class of random graph models and a very general definition of what makes the v.o.i.\@ interesting.  
In this paper, we further develop the ideas in \cite{lyzinski2017consistent}, with the aim of developing a theoretical regime in which to ground the notion of adversarial contamination in VN. In addition, their results are derived in the setting of a single vertex of interest; since many real application problems involve finding similar groups of nodes, we extend their results to multiple vertices of interest.

There has been significant recent attention towards better understanding the impact of adversarial attacks on machine learning methodologies (see, for example, \cite{huang2011adversarial,cai2015robust,papernot2016limitations,adv1,adv2}). 
Herein, we define an adversarial attack on a machine learning algorithm to be a mechanism that changes the data distribution in order to negatively affect algorithmic performance; see Definition \ref{def:Adv}. 
From a practical standpoint, adversarial attacks model the very real problem of having data compromised; 
if an intelligent agent has access to the data and algorithm, the agent may want to modify the data or the algorithm to give the wrong prediction/inferential conclusion.  
Although there has been much work on adversarial modeling in machine learning, there has been less theory developed for adversarial attacks from a statistical perspective.  

The adversarial framework we consider is similar to the model considered in \cite{cai2015robust}, and it is motivated by the example in the previous section in which the addition of the vertices without correspondences to $G_2$ negatively impacted VN performance.
Suppose that we are interested in performing vertex nomination on a graph pair, 
but an adversary randomly adds and deletes some edges and/or vertices in the second graph.  
For example, suppose we are trying to find influencers on Instagram by vertex matching to Facebook.
An influencer that has knowledge of our procedure may attempt to make our algorithm fail in its nominations, perhaps by friending and de-friending people on Facebook.  
Even if our vertex nomination scheme was working well prior to encountering the adversary, it may not be after modification by the adversary.  

From a statistical standpoint, what can we say about the statistical consistency of our original vertex nomination rule?  
Our motivating example suggests that there are adversaries that can render our vertex nomination scheme no longer consistent, but theory is needed both to explain why that may be the case and to properly frame the problem.  Hence, to answer these questions, we further develop the theory in \cite{lyzinski2017consistent} to situate the notion of adversarial contamination within the idea of maximal consistency classes for a given VN rule (Section \ref{sec:CC}).
In this framework, the goal of an adversary is to move a model out of a rule's consistency class. 
We demonstrate with real and synthetic data examples how an adversary is able to move a model out of a rule's consistency class.  We finish with a brief discussion on how regularization can effectively recover consistency, though we leave this for future work.

\vspace{3mm}
\noindent{\bf Notation:} 
See Table \ref{table:1} for frequently used notation.

\begin{table}[h!]
\centering
 \begin{tabular}{|l | l|} 
 \hline
 Notation & Description \\ [0.5ex] 
 \hline
 &\\
 $[k]$ & The set of integers $\{1,2,3,\ldots,k\}$\\
 &\\
 $G = (V,E)$ \ \ \ \ \ & A (random) graph with vertex set $V$ and edge set $E$\\
 &\\
 $G_1 = (V_1, E_1)$, $G_2 = (V_2, E_2)$ & Two random graphs with a presumed shared set of vertices \\
  &\\
$C$ & A core set of vertices shared between two graphs \\
&\\
$J_1$, $J_2$ & Junk vertices not shared between graphs \\
&\\
$\mathcal{G}_n$  & The set of $n$-vertex labeled graphs \\
    &\\
\multirow{2}{2em}{$F^{(n,m)}_{c,\theta}$} & A nominatable distribution on $\mathcal{G}_n \times  \mathcal{G}_m$ \\ &with $c$ shared vertices and parameter $\theta$ \\
 &\\
 $\mathcal{N}_{n,m}$ & The set of nominatable distributions on $\mathcal{G}_n \times \mathcal{G}_m$ \\
 &\\
 $g, g_1, g_2$ & Observed graphs \\
  &\\
 $V^*$ & A vertex set of interest shared between two graphs \\
  &\\
 $v^*$ & A single vertex of interest \\
  &\\
 $\mo$ & An obfuscation function changing observed vertex labels \\
  &\\
  $\mO_W$ & The set of obfuscating functions mapping a vertex set to $W$
  \\ &\\
  
 $\mathcal{T}_W$ & The set of total orderings of the elements of a set $W$ \\
  &\\
  $\mathcal{I}(u; g)$ & The set of vertices in $g$ topologically equivalent to $u$ \\
  &\\
 \multirow{2}{2em}{$\Phi(g_1, \mo( g_2), V^*)$} & A vertex nomination scheme with vertex set of interest \\ & $V^*$ and observed graphs $g_1$ and $\mo(g_2)$ \\ 
 &\\
$\mathfrak{r}_{\Phi}(g_1,g_2,\mo,V^*,S)$ & The set of ranks of a set $S$ under $\Phi(g_1, \mo(g_2), V^*)$ \\
&\\
\hline
\end{tabular}
\caption{Table of frequently used notation}
\label{table:1}
\end{table}

\section{Vertex Nomination and Consistency}
\label{sec:VN}
Before discussing how to define adversarial attacks, we discuss the previous work of \cite{lyzinski2017consistent}, the first of its kind to derive the Bayes Optimal vertex nomination scheme for one vertex. This work can be viewed as a follow-on of that work, in which we provide a groundwork for the rigorous framing of an adversary in vertex nomination.

First, we will situate our analysis of the VN problem in the very general framework of nominatable distributions.
\definition[Nominatable Distribution]
For a given $n,m\in\mathbb{Z}>0$, the set of \emph{Nominatable Distributions of order $(n,m)$}, denoted $\mathcal{N}_{n,m}$, is the collection of all families of distributions  $\mathbf{F}^{(n,m)}_{\Theta}$  of the following form 
$$\{F^{(n,m)}_{c,\theta} \ \text{ s.t. $0\leq c\leq\min(n,m) \in \mathbb{Z}, \theta \in \Theta\subset \mathbb{R}^{d(n,m)}$}\}$$ where $F^{(n,m)}_{c,\theta}$ is a distribution on $\mathcal{G}_n \times \mathcal{G}_m$ parameterized by $\theta \in \Theta$ satisfying: 
\begin{enumerate}
\item{The vertex sets $V_1 = \{v_1,v_2,...,v_n\}$ and 
$V_2 = \{u_1,u_2,...,u_m\}$ satisfy $v_i = u_i$ for $0 < i \leq c$. 
We refer to $C = \{v_1,v_2,...,v_c\} = \{u_1,u_2,...,u_c\}$ as the core vertices. These are the vertices that are shared across the two graphs and imbue the model with a natural notion of corresponding vertices.}
\item{Vertices in $J_1 = V_1 \setminus C$ and $J_2 = V_2 \setminus C$, satisfy $J_1 \cap J_2 = \emptyset$. We refer to $J_1$ and $J_2$ as junk vertices. These are the vertices in each graph that have no corresponding vertex in the other graph}
\item{The induced subgraphs $G_1[J_1]$ and $G_2[J_2]$ are conditionally independent given $\theta$.}
\end{enumerate}

\noindent The vertices in $C$ are those that have a corresponding paired vertex in each graph; where corresponding can be defined very generally.
Corresponding vertices need not correspond to the same person/user/account, rather corresponding vertices are understood as those that share a desired property (for example, a role in the network) across graphs.
In particular, we will assume that the vertices of interest in $G_1$ \vv{have corresponding vertices in $G_2$, and that these corresponding vertices are} the vertices of interest in $G_2$.

Having access to the vertex labels would then render the VN problem trivial.
To model the uncertainty often present in data applications, where the vertex labels (or correspondences) are unknown a priori we adopt the notion of \textit{obfuscation functions} from~\cite{lyzinski2017consistent}.
\definition[Obfuscating Function] Let $(G_1,G_2) \sim F^{(n,m)}_{c,\theta} \in \mathcal{N}_{n,m}$, and let $W$ be a set satisfying $W \cap V_i = \emptyset$ for $i = 1,2.$
An obfuscating function $\mo : V_2 \mapsto W$ is a bijection from $V_2$ to $W$.
We refer to $W$ as an obfuscating set, and we let $\mO_W$ be the set of all such obfuscation functions.

\subsection{\vv{VN in the Setting of a Single Vertex of Interest}}
\label{sec:VN_multiple}

With these two definitions in place, we \vv{now} present the definition of a vertex nomination scheme for a single vertex of interest as in \cite{lyzinski2017consistent}.  
\vv{In Section \ref{sec:VN_multiple}, we will extend the definition of a vertex nomination scheme to encompass multiple vertices of interest.  In the remainder of this section, we will l}et $v^* \in V_1$ be the \vv{given} vertex of interest \vv{in $G_1$}.

\begin{definition} \label{def:VN_single}
(VN Scheme for single VOI)
Let $n,m\in\mathbb{Z}>0$, and for each $g\in\mathcal{G}_m$, $u\in V(g)$, let
\begin{align*}
\mathcal{I}(u;g)=\{w\in V(g)\text{ s.t. } &
	\exists\text{ an automorphism }\sigma\text{ of }g,\text{ s.t. }\,\sigma(u)=w\}.
\end{align*}
Let $W$ be an obfuscating set and $\mo\in\mathfrak{O}_W$ be given.
For a set $A$, let $\calT_A$ denote the set of all total orderings of the elements of $A$.
A \emph{vertex nomination scheme} is a function 
$\Phi: \gn \times \mo(\gm) \times V_1 \rightarrow \calT_{W}$ satisfying the following consistency property:
If for each $u\in V_2$, we define $\text{rank}_{\Phi(g_1,\mo(g_2),v^*)}\big(\mo(u)\big)$ to be the position of $\mo(u)$ in the total ordering provided by $\Phi(g_1,\mo(g_2),v^*)$, and we define
$\mathfrak{r}_{\Phi}:\gn\times\gm\times\mathfrak{O}_W\times V_1 \times2^{V_2}\mapsto 2^{[m]}$ via
$$\mathfrak{r}_{\Phi}(g_1,g_2,\mo,v^*,S)=\{\text{rank}_{\Phi(g_1,\mo(g_2),v^*)}\big(\mo(u)\big)\text{ s.t. }u\in S \},$$
then we require that for any $g_1\in\gn,$ $g_2\in\gm$, $v^*\subset V_1$, obfuscating functions $\mo_1,\mo_2\in\mathfrak{O}_W$ and any $u\in V(g_2)$, 
\begin{align}
\label{eq:consis2}
&\mathfrak{r}_{\Phi}(g_1,g_2,\mo_1,v^*,\mathcal{I}(u;g_2))=\mathfrak{r}_{\Phi}(g_1,g_2,\mo_2,v^*,\mathcal{I}(u;g_2))\\
\notag &\hspace{10mm}\Longleftrightarrow \\
\notag &\mo_2\circ\mo_1^{-1}\big( \mathcal{I}(\Phi(g_1,\mo_1(g_2),V^*)[k]);\mo_1(g_2)\big)  =\mathcal{I}\left( \Phi(g_1,\mo_2(g_2),v^*)[k];\mo_2(g_2) \right)\\
&\hspace{20mm} \text{ for all }k\in[m],\notag
\end{align}
where $\Phi(g_1,\mo(g_2),v^*)[k]$ denotes the $k$-th element (i.e., the rank-$k$ vertex) in the ordering $\Phi(g_1,\mo(g_2),v^*)$.
We let $\mathcal{V}_{nm}$ denote the set of all such VN schemes.
\end{definition}

\begin{remark}
The consistency criterion, Eq. \ref{eq:consis2}, models the property that a sensibly-defined vertex nomination scheme should view all vertices in a given $\mathcal{I}_g(u)$ as being equally ``interesting'' in $G_2$.
These vertices are topologically indistinguishable, and thus are only separated by their labels which have been obfuscated via $\mo$.
Truly obfuscated vertex labels should be independent of the obfuscation function, and the consistency criterion requires that the set of ranks of each set of equivalent vertices (i.e., each $\mathcal{I}_{g_2}(u)$) does not depend on the particular choice of obfuscation function.
\end{remark}

One can already begin to see how one might extend these definitions to multiple vertices of interest; note that $\Phi$ is a function of two graphs and a single vertex.  It will be natural to require $\Phi$ to be a function of two graphs and a vertex \emph{set} instead.  We give these definitions in Section \ref{sec:VN_multiple}.  We first define the error for the vertex nomination scheme \vv{defined above}.

\begin{definition}[VN loss function, level-$k$ error for single VOI] Let $\Phi$ be a vertex nomination scheme, and $\mathfrak{o}$ an obfuscating function.  For $(g_1, g_2)$ realized from $(G_1, G_2)\sim F_{c,n,m,\theta}$ with vertex of interest $v^* \in C$, and $k \in [m-1]$, we define the level-$k$ nomination loss via
\begin{align*}
    \ell_{k}\left(\Phi, g_{1}, g_{2}, v^{*}\right) &=\mathds{1}\big\{\text{rank}_{\Phi\left(g_{1},\mathfrak{o} \left(g_{2}\right), v^{*}\right)}\left(\mathfrak{o}\left(v^{*}\right)\right) \geq k+1\big\}, \\
    &= 1-\mathds{1}\left\{\text{rank}_{\Phi\left(g_{1}, \mathfrak{o}\left(g_{2}\right), v^{*}\right)}\left(\mathfrak{o}\left(v^{*}\right)\right) \leq k\right\}.
\end{align*}
The level $k$ error of $\Phi$ at $v^*$ is then defined to be
\begin{align*}
    L_{k}\left(\Phi, v^{*}\right)&=\mathbb{E}_{\left(G_{1}, G_{2}\right) \sim F_{c, n, m, \theta}}\left[\ell_{k}\left(\Phi, G_{1}, G_{2}, v^{*}\right)\right] \\
    &= \mathbb{P}_{\left(G_{1}, G_{2}\right) \sim F_{c, n, m, \theta}}\left[\operatorname{rank}_{\Phi\left(G_{1}, \mathfrak{o}\left(G_{2}\right), v^{*}\right)}\left(\mathfrak{o}\left(v^{*}\right)\right) \geq k+1\right].
\end{align*}
\end{definition}

The level $k$ error is simply the probability that the rank of the vertex of interest in $g_2$ is not in the nomination list; this matches our intuition for what the error should be. To discuss the notion of consistency, we need to assume that the core set $C$ of the nominated are \emph{nested} in the following sense. 

\begin{definition}[Nested Cores]
Let $\mathbf{F}=\left(F^{(n,m_n)}_{c_n,\theta_n}\right)_{n=n_0}^\infty$ be a sequence of distributions in $\mathcal{N}$.
We say that $\bf F$ has {\em nested cores} if there exists an $n_1$ such that for all $n_1\leq n< n'$, if $(G_1,G_2)\sim F^{(n,m_n)}_{c_n,\theta_n}$ and $(G_1',G_2')\sim F^{(n',m_{n'})}_{c_{n'},\theta_{n'}}$, we have,
letting $C$ and $C'$ be the core vertices
associated with $F^{(n,m_n)}_{c_n,\theta_n}$
and $F^{(n',m_{n'})}_{c_{n'},\theta_{n'}}$ respectively,
and denoting the junk vertices $J_1,J_1',J_2,J_2'$
analogously,
\begin{itemize}
\item[i.] $V(G_1)=C\cup J_1\subset V(G_1')=C'\cup J_1'$;
\item[ii.] $V(G_2)=C\cup J_2\subset V(G_2')=C'\cup J_2'$;
\item[iii.] $C\subset C'$.
\end{itemize}
\end{definition}

\vv{In \cite{lyzinski2017consistent}, for any given nominatable distribution $F_{c,\theta}^{n,m}$, a Bayes optimal VN scheme is defined that is simultaneously optimal at all levels $k$.
We will denote this optimal scheme via $\Phi^*=\Phi^*_{F_{c,\theta}^{n,m}}$, and its associated level $k$ loss via $L^*_k$.
The notion of consistency in VN is then defined as follows.}
\vv{\definition[Level $k_n$ Consistent VN Rules in the single v.o.i. setting] Let ${\bf F} =(F^{(n,m_n)}_{c_n,\theta_n})_{n=n_0}^{n=\infty}$
be a sequence of nominatable distributions in $\mathcal{N}$ with nested cores satisfying $\lim_{n\to \infty} m_n = \infty$. 
For a given non-decreasing sequence $(k_n)$, we say that a VN rule $\mathbf{\Phi} = (\Phi_{n,m_n})_{n=n_0}^{n=\infty}$ is 
level-$(k_n)$ consistent for vertex of interest $v^*\in C_{1}$ with respect to ${\bf F}$ if 
$$\lim_{n\to \infty} L_{k_n}(\Phi_{n,m_n},v^*) - L^{*}_{k_n}(v^*) = 0,$$}

\vv{We say that a VN rule $\mathbf{\Phi}$ is \emph{universally level-$(k_n)$ consistent} if it is level-$(k_n)$  consistent for all nested-core nominatable sequences $\bf F$.}
Before \vv{presenting vertex nomination schemes in the multiple v.o.i. setting, we first} present an important consistency result given in \cite{lyzinski2017consistent}, which says that there are \vv{{\bf no}} universally consistent vertex nomination schemes.  

\begin{theorem}[Corollary 28 of \cite{lyzinski2017consistent}] \label{thm:noconsistent} Let $\varepsilon \in (0,1)$ be arbitrary, and consider a VN rule $\mathbf{\Phi} = (\Phi_{n,m})$.  For any nondecreasing sequence $(k_n)_{n=n_0}^{\infty}$ satisfying $k_n = o(m)$, there exists a sequence of distributions $F_{c,n,m,\theta}$ in $\mathcal{N}$ with nested cores such that
$$\lim \sup _{n \rightarrow \infty} L_{k_{n}}^{*}\left(v^{*}\right)=\epsilon<1=\lim _{n \rightarrow \infty} L_{k_{n}}\left(\Phi_{n, m}, v^{*}\right).$$
\end{theorem}

This result is markedly different from the \vv{setting of classical classification}, in which there exist universally consistent classifiers.  
\vv{In Section \ref{sec:adv}, we will explore the ramifications of Theorem \ref{thm:noconsistent} on our understanding of adversarial attacks on VN rules; effectively} such a result might mean that an adversary \vv{acts by moving a given distribution outside of the ``consistency class'' of a given nomination rule (see Section \ref{sec:CC} for detail).} 

We next extend definitions to the more practical setting of multiple vertices of interest.

\subsection{Extension to Multiple Vertices of Interest}
\label{sec:VN_multiple}

We will now rigorously define the VN problem and consistency within the VN framework for multiple vertices of interest.  
Combined with the results on consistency classes in Section \ref{sec:CC}, this will allow us to provide a statistical basis for understanding adversarial attacks in VN.  Our definitions and notation are based on those in the previous section, though we have a few more general requirements.  Recall that \cite{lyzinski2017consistent} defined a vertex nomination scheme as a function from $\Phi: \gn \times \mo(\gm) \times V_1 \rightarrow \calT_{W}$ satisfying a certain consistency property.  The extension to multiple vertices of interest requires that $\Phi$ be a function taking in a \emph{set} of vertices.  The rigorous definition is given below.



\begin{definition} \label{def:VN}
(VN Scheme)
Let $n,m\in\mathbb{Z}>0$, and for each $g\in\mathcal{G}_m$, $u\in V(g)$, and again let
\begin{align*}
\mathcal{I}(u;g)=\{w\in V(g)\text{ s.t. } &
	\exists\text{ an automorphism }\sigma\text{ of }g,\text{ s.t. }\,\sigma(u)=w\}.
\end{align*}
Let $W$ be an obfuscating set and $\mo\in\mathfrak{O}_W$ be given.
For a set $A$, let $\calT_A$ denote the set of all total orderings of the elements of $A$.
A \emph{vertex nomination scheme} is a function 
$\Phi: \gn \times \mo(\gm) \times 2^{V_1} \rightarrow \calT_{W}$ satisfying the following consistency property:
If for each $u\in V_2$, we define $\text{rank}_{\Phi(g_1,\mo(g_2),V^*)}\big(\mo(u)\big)$ to be the position of $\mo(u)$ in the total ordering provided by $\Phi(g_1,\mo(g_2),V^*)$, and we define
$\mathfrak{r}_{\Phi}:\gn\times\gm\times\mathfrak{O}_W\times 2^{V_1}\times2^{V_2}\mapsto 2^{[m]}$ via
$$\mathfrak{r}_{\Phi}(g_1,g_2,\mo,V^*,S)=\{\text{rank}_{\Phi(g_1,\mo(g_2),V^*)}\big(\mo(u)\big)\text{ s.t. }u\in S \},$$
then we require that for any $g_1\in\gn,$ $g_2\in\gm$, $V^*\subset V_1$, obfuscating functions $\mo_1,\mo_2\in\mathfrak{O}_W$ and any $u\in V(g_2)$, 
\begin{align}
\label{eq:consis}
\mathfrak{r}_{\Phi}(g_1,g_2,\mo_1,V^*,\mathcal{I}(u;g_2))&=\mathfrak{r}_{\Phi}(g_1,g_2,\mo_2,V^*,\mathcal{I}(u;g_2))\\
\notag &\Leftrightarrow \\
\notag \mo_2\circ\mo_1^{-1}\big( \mathcal{I}(\Phi(g_1,\mo_1(g_2),V^*)[k]);\mo_1(g_2)\big) & =\mathcal{I}\left( \Phi(g_1,\mo_2(g_2),V^*)[k];\mo_2(g_2) \right)\\
&\hspace{20mm} \text{ for all }k\in[m],\notag
\end{align}
where $\Phi(g_1,\mo(g_2),V^*)[k]$ denotes the $k$-th element (i.e., the rank-$k$ vertex) in the ordering $\Phi(g_1,\mo(g_2),V^*)$.
We let $\mathcal{V}_{nm}$ denote the set of all such VN schemes.
\end{definition}


A VN scheme is an information retrieval tool for efficiently querying large network data sets.
Rather than naively searching $G_2$ for interesting vertices, an appropriate VN scheme provides a rank list of the vertices in $G_2$ that, ideally, allows users to identify v.o.i.\@ in $G_2$ in a time-efficient manner.
As such, to measure the performance of a VN scheme on multiple vertices, we will adopt a recall-at-k/precision-at-k framework.  More precisely, we have the following definition.

\definition[Level $k$ Nomination Loss] Let $\Phi \in \mathcal{V}_{n,m}$ be a vertex nomination scheme, $W$ an obfuscating set, and $\mo\in\mathfrak{O}_W$.  
Let $(g_1,g_2)$ be realized from $(G_1, G_2) \sim F^{(n,m)}_{c\theta}\in \mathcal{N}_{n,m}$ with a vertex of interest set $V^* \subset C$.
For $k \in [m-1]$, we define the level-k nomination losses via
\begin{align*}
\ell^{(1)}_k (\Phi,g_1,g_2,V^*) :&= \tfrac{\sum_{v \in V^*} \mathds{1}\{ \text{rank}_{\Phi(g_1, \mo(g_2), V^*)}( \mo(v)) \geq k + 1\}}{|V^*|} \\
&= 1 - \tfrac{\sum_{v \in V^*} \mathds{1}\{ \text{rank}_{\Phi(g_1, \mo(g_2), V^*)}( \mo(v)) \leq k\}}{|V^*|}\\
\ell^{(2)}_k (\Phi,g_1,g_2,V^*) :&=\vv{1- \tfrac{\sum_{v \in V^*} \mathds{1}\{ \text{rank}_{\Phi(g_1, \mo(g_2), V^*)}( \mo(v)) \leq k\}}{|k|},}
\end{align*}
where the $(1)$ and $(2)$ superscripts refer to recall and precision respectively.  The error of a VN scheme is then defined as the expected loss.  
To wit, we have the following definition.
\definition[Level-$k$ Error]
Let $\Phi \in \mathcal{V}_{n,m}$ be a vertex nomination scheme, $W$ an obfuscating set, and $\mo\in\mathfrak{O}_W$.
The level-$k$ error of $\Phi$ for $V^*\subset C$ and $F^{(n,m)}_{c,\theta}\in\mathcal{N}$ is defined as 
\begin{align*}
L^{(1)}_k (\Phi, V^*) :&= \e_{(G_1, G_2) \sim F^{(n,m)}_{c,\theta}}[ \ell^{(1)}_k (\Phi, G_1, G_2, V^*)] \\
&= \frac{1}{|V^*|} \sum_{v \in V^*}
\p_{F^{(n,m)}_{c,\theta}}
\bigg( \text{rank}_{\Phi(G_1, \mo(G_2),V^*)}(\mo(v)) \geq k + 1 \bigg)\\
L^{(2)}_k (\Phi, V^*) :&= \e_{(G_1, G_2) \sim F^{(n,m)}_{c,\theta}}[ \ell^{(2)}_k (\Phi, G_1, G_2, V^*)] \\
&=\vv{1- \frac{1}{|k|} \sum_{v \in V^*}
\p_{F^{(n,m)}_{c,\theta}}
\bigg( \text{rank}_{\Phi(G_1, \mo(G_2),V^*)}(\mo(v)) \leq k  \bigg)}
\end{align*}
The \emph{level-k Bayes optimal scheme} is defined as any element
\begin{align*}
\Phi^*_{k,V^*}&\in\text{argmin}_{\Phi\in\mathcal{V}_{nm}}L^{(1)}_k (\Phi, V^*)=\text{argmin}_{\Phi\in\mathcal{V}_{nm}}L^{(2)}_k (\Phi, V^*),  \end{align*}
with corresponding errors $L^{*,(1)}_k$ and $L^{*,(2)}_k$.

In the \vv{almost sure} absence of symmetries amongst the vertices in $V^*$ (i.e., $\mathcal{I}(v,G_2)=\{v\}$ for all $v\in V^*$), the derivation of the Bayes optimal scheme in the present $|V^*|>1$ setting mimics that of the $|V^*|=1$ setting presented in \cite{lyzinski2017consistent}.

\subsubsection{Bayes Optimal VN Scheme Construction}
\label{sec:BOcon}
\vv{With notation as above, 
Let $n,m$ be fixed and let $V^*\subset V_1\cap V_2$ be fixed.
Let $W$ be an obfuscating set and $\mo\in\mathfrak{O}_W$.
Further assume that $F=F^{(n,m)}_{c,\theta}$ is such that $\mathcal{I}(v,G_2)\stackrel{a.s.}{=}\{v\}$ for all $v\in V^*$, so that $F$ is supported on 
$$\mathcal{G}_{n,m}^a:=
\left\{(g_1,g_2)\in\gn\times\gm\text{ s.t. }\mathcal{I}(v;g_2)=\{v\}\text{ for all }v\in V^* \right\}.$$
For each $(g_1,g_2)\in \gnma$ define
\begin{align*}
(g_1, [\mo(g_2)])&= \bigg\{ (g_1, \tilde g_2) \in \gnma: \mo(\tilde g_2) \simeq \mo(g_2) \bigg\} \\
&=\bigg\{  (g_1, \tilde g_2) \in \gnma: \tilde g_2 \simeq g_2 \bigg\}. 
\end{align*}
where $\simeq$ denotes graph isomorphism.
For each $w \in W$ and $u \in V_2$, we also define the following restriction
\begin{align*}
(g_1, [\mo(g_2)])_{w = \mo(u)} &=\bigg\{ (g_1, \tilde g_2) \in \gnma\text{ s.t. } \mo(\tilde g_2) = \sigma( \mo(g_2)),\\& \ \ \ \ \ \ \ \ \ \ \ \ \ \ \ \text{         $\sigma$ an isomorphism, } \sigma(w)=\mo(u) \bigg\} \\
&=\bigg\{  (g_1, \tilde g_2) \in \gnma\text{ s.t. } \tilde g_2 = \sigma( g_2), \\& \ \ \ \ \ \ \ \ \ \ \ \ \  \text{ $\sigma$ an isomorphism, } \sigma(\mo^{-1}(w))= u\bigg\},
\end{align*}
and for $S\subset V_2,$ define 
$$(g_1, [\mo(g_2)])_{w \in \mo(S)}=\bigcup_{u\in S}\,\,(g_1, [\mo(g_2)])_{w = \mo(u)}.$$
Choose graphs
\begin{equation}
\label{eq:g}
\mathbf{g}=\left\{\left(g_1^{(i)},g_2^{(i)}\right)\right\}_{i=1}^h
\end{equation} 
so that the sets 
$$\left\{\left(g_1^{(i)},[\mo(g_2^{(i)})]\right)\right\}_{i=1}^h
$$
 partition $\gnma$.
To ease notation, we will denote this partition via $\mathcal{P}^{\bf g}_{n,m}$.
We will next define a Bayes optimal scheme $\Phi^*$ (optimal under both loss functions simultaneously for all $k\in[m-1]$ for the above $F$ supported on $\gnma$).  }

\vv{For ease of notation, for each $i\in[h]$ and $u\in W$, define $$P_u^i := \PF\bigg( (g_1^{(i)}, [\mo(g_2^{(i)})])_{u\in \mo(V^*) }\,\, \big|\,\, (g_1^{(i)}, [\mo(g_2^{(i)})]) \bigg)$$
Then, set (where ties are broken in a fixed but arbitrary manner)
\begin{align*}
    \Phi^*(g_1^{(i)} , \mo(g_2^{(i)}), V^*)[1] &\in \argmax_{\substack{u\in W }}  \,\,P_u^i \\
\Phi^*(g_1^{(i)} , \mo(g_2^{(i)}), V^*)[2] &\in \argmax_{\substack{u\in W\setminus\{ \Phi^*[1]\}}}  P_u^i \\
&\vdots \\
\Phi^*(g_1^{(i)} , \mo(g_2^{(i)}), V^*)[m] &\in \argmax_{\substack{u\in W\setminus\{\cup_{j< m}\{\Phi^*[j]\}}}  P_u^i.
\end{align*}
For each element 
$$(g_1,g_2)\in( g^{(i)}_1,[\mo(\tilde g_2^{(i)})])\setminus\{(g_1^{(i)},g_2^{(i)})\},$$ choose an isomorphism $\sigma$ such that $\mo(g_2)=\sigma(\mo(g_2^{(i)}))$, and define
$${\Phi^*}(g_1,\mo(g_2),V^*)=\sigma({\Phi^*}(g_1^{(i)},\mo(g_2^{(i)}),V^*)).$$
See Appendix \ref{sec:BO} 
for a proof of the optimality of such a scheme.}

Bayes optimal schemes when symmetries exist for the v.o.i.---i.e., when there are $v\in V^*$ such that $|\mathcal{I}(v,;g_2)|>1$---offer additional complications and, in the case when $|V^*|=1$ done in \cite{lyzinski2017consistent}, little additional insight. 
Precisely defining the Bayes optimal scheme in the case of symmetries when $|V^*|>1$ is notationally and technically nontrivial, and is the subject of current research.

\subsubsection{Consistency in VN with $|V^*|>1$}
\label{sec:conVN}
Consistency in the VN framework for multiple vertices is then defined as follows.

\definition[Level $k_n$ Consistent VN Rules] Let ${\bf F} =(F^{(n,m_n)}_{c_n,\theta_n})_{n=n_0}^{n=\infty}$
be a sequence of nominatable distributions in $\mathcal{N}$ with nested cores satisfying $$\lim_{n\to \infty} m_n = \infty.$$ 
For a given non-decreasing sequence $(k_n)$, we say that a VN rule $\mathbf{\Phi} = (\Phi_{n,m_n})_{n=n_0}^{n=\infty}$ is (where \vv{the level $k_n$-losses here are computed with respect to $F_n=F^{(n,m_n)}_{c_n,\theta_n}$})
\begin{itemize}
\item [i.] level-$(k_n)$ recall consistent for nested $V^*_{n}\in C_{n}$ with respect to ${\bf F}$ if 
$$\lim_{n\to \infty} L^{(1)}_{k_n}(\Phi_{n,m_n},V^*_n) - L^{*,(1)}_{k_n}(V^*_n) = 0,$$
for any sequence of obfuscating functions of $V_2$ with $|V_2| = m_n$. 
Note that the level $k_n$-loss here is computed with respect to $F_n=F^{(n,m_n)}_{c_n,\theta_n}$.
\item[ii.] level-$(k_n)$ precision consistent for for nested $V^*_{n}\in C_{n}$ with respect to ${\bf F}$ if 
$$\lim_{n\to \infty} L^{(2)}_{k_n}(\Phi_{n,m_n},V^*_n) - L^{*,(2)}_{k_n}(V^*_n) = 0,$$
for any sequence of obfuscating functions of $V_2$ with $|V_2| = m_n$. 
\end{itemize}

We say that a VN rule $\mathbf{\Phi}$ is \emph{universally level-$(k_n)$ $\binom{\text{precision}}{\text{recall}}$ consistent} if it is level-$(k_n)$ $\binom{\text{precision}}{\text{recall}}$ consistent for all nested-core nominatable sequences $\bf F$.
Theorem \ref{thm:noconsistent} in the previous section (Corollary 28 from \cite{lyzinski2017consistent}) proves that universally consistent VN schemes do not exist for any nondecreasing integral sequences $(k_n)$ satisfying $k_n=o(m_n)$ and any $(V^*_n)$ satisfying $|V^*_n|=\Theta(1)$. 
Beyond the ramifications for practically implementing VN in streaming or evolving network environments considered in \cite{lyzinski2017consistent}, this lack of universal consistency is also the motivating result for our statistical approach to adversarial contamination in VN.
Indeed, a simple consequence of the lack of universal consistency is that for any VN rule there are nominatable sequences for which the rule is not consistent.
An adversary could then be understood as a probabilistic mechanism designed to transform nominatable sequences for which the rule is consistent into nominatable sequences for which the rule is not consistent.

To develop this reasoning further, we next develop the notion of (maximal) consistency classes in the VN framework.

\subsection{VN Consistency Classes}
\label{sec:CC}

We next explore the concept of consistency classes in VN, with an eye towards the development of a statistical adversarial contamination framework for VN.
First, let $\mathfrak{N}_{\bf V^*}$ be the collection of all nested-core nominatable sequences with nested v.o.i. ${\bf V}^*=(V^*_{n}\subset C_n)$. 
For a given VN rule $\mathbf{\Phi}$, v.o.i. sequence ${\bf V}^*$ satisfying $|V^*_n|=\Theta(1)$, and nondecreasing sequence $(k_n)$ (satisfying the growth condition \vv{$k_n=o(n)$} of Theorem \ref{lem:CC}), the level-$(k_n)$ $\binom{\text{precision}}{\text{recall}}$ consistency class of $\mathbf{\Phi}$ is defined to be
\begin{align*}
\mathfrak{C}_{\mathbf{\Phi}}^{(k_n)}=\bigg\{\mathbf{F}\in\mathfrak{N}_{\mathbf{V}^*}\text{ s.t. }\mathbf{\Phi}&\text{ is level-}(k_n)\,\,\binom{\text{precision}}{\text{recall}}\text{ consistent for }\mathbf{F}  \bigg\}.
\end{align*}
The lack of universal consistency ensures that $\mathfrak{C}_{\mathbf{\Phi}}^{(k_n)}\neq \mathfrak{N}_{\mathbf{V}^*}$ for any rule $\mathbf{\Phi}.$

It is natural to ask if there are a finite number of VN rules $\{\mathbf{\Phi}_i\}$ such that $\cup_i \mathfrak{C}_{\mathbf{\Phi}_i}^{(k_n)}=\mathfrak{N}_{\mathbf{V}^*}$.  
An affirmative answer would allow for ensemble methods to practically overcome the lack of universally consistent rules, and hence practically overcome any adversarial attack in the VN framework.
We will see in Section \ref{sec:ccc} that the answer is, as expected, no, and any partition of $\mathfrak{N}_{\mathbf{V}^*}$ into maximal consistency classes necessarily contains infinite parts; see Theorem \ref{lem:CC}.
As a consequence, ensemble methods cannot recover universal consistency in VN.
The insights developed in Section \ref{sec:ccc} further motivate the development of adversarial contamination regimes for a given rule $\mathbf{\Phi}$.
The idea behind adversarial contamination is simple in this framework: 
the adversary contaminates elements $\bf F\in\mathfrak{C}_{\mathbf{\Phi}}^{(k_n)}$ transforming them into $\bf F'\in\mathfrak{N}_{\mathbf{V}^*}\setminus \mathfrak{C}_{\mathbf{\Phi}}^{(k_n)}$.

\subsubsection{Counting Consistency Classes}
\label{sec:ccc}

How can a practitioner mitigate the impact of a lack of universal consistency?
One idea would be to consider ensemble methods, as the practical implications of the lack of universal consistency can be mitigated if universally consistent ensemble schemes exist.
In this section, we will formalize the notion of maximal VN consistency classes and prove that infinitely many maximal consistency classes exist.
We begin with defining the notion of maximal consistency classes in the VN-framework. 
\begin{definition}[Maximal Consistency Class]
As above, let $\mathfrak{N}_{\bf V^*}$ be the collection of all nested-core nominatable sequences with nested v.o.i. ${\bf V}^*=(V^*_{n}\subset C_n)$.
For a nondecreasing integer sequence $(k_n)$, we say that $\mathfrak{C}\in \mathfrak{N}_{\bf V^*}$ is a maximal level-$(k_n)$ $\rp$ consistency class for ${\bf V}^*$ if the following two conditions hold.
\begin{itemize}
\item[i.] There exists a VN rule $\Phi$ that is jointly level-$(k_n)$ $\binom{\text{precision}}{\text{recall}}$ consistent for ${\bf V}^*$ for each $\mathbf{F}\in\mathfrak{C}$;
\item[ii.] If $\mathbf{F}'\notin\mathfrak{C}$, then there does not exist a VN rule $\Phi$ that is jointly level-$(k_n)$ $\rp$ consistent for ${\bf V}^*$ for each $\mathbf{F}\in\mathfrak{C}\cup\{\mathbf{F}'\}$.
\end{itemize}
\end{definition} 
\noindent
A natural question to ask is whether it is possible to partition $\nn$ into a finite number of maximal level-$(k_n)$ consistency classes for a particular sequence $(k_n)_{n=1}^\infty$?
Our next result---Theorem \ref{lem:CC}---shows that for any integer sequence $(k_n)$ satisfying a modest growth condition, any partition of $\mathfrak{N}$ into maximal level-$(k_n)$ consistency classes must include at least countably infinite parts, thus erasing the hope that ensemble methods can recover universal consistency and practically mitigate the effect of any VN adversarial attack.

\begin{theorem}
\label{lem:CC}
Let $(k_n)$ be a sequence of nondecreasing integers satisfying $k_n=o(n)$, and let ${\bf V}^*$ be a nested sequence of vertices of interest satisfying $|V^*_n|=\Theta(1)$. 
\begin{itemize}
\item[i.] Let $\nn=\cup_{\alpha\in\mathcal{A}}\mathfrak{C}_\alpha$ be a partition of $\nn$ into maximal level-$(k_n)$ recall consistency classes, then $|\mathcal{A}|=\infty$.
\item[ii.] Let $\nn=\cup_{\alpha\in\mathcal{A}}\mathfrak{C}_\alpha$ be a partition of $\nn$ into maximal level-$(k_n)$ precision consistency classes.
If $k_n=\Theta(1)$, then $|\mathcal{A}|=\infty$.
\end{itemize}
\end{theorem}

\noindent The proof of this Theorem can be found in Appendix \ref{app:CC}.


\subsubsection{Verification functions}
\label{sec:verify}
In the presence of an adversarial attack, is it possible to, without additional supervision, verify if a given VN scheme is working on a given $F_{c,\theta}^{(n,m)}\in\mathcal{N}_{n,m}$? 
In other words, given a nondecreasing integer sequence $(k_n)$, $(g_1,g_2)\in\gn\times\gm$, and v.o.i. $V^*_n$, can we consistently estimate the \emph{verification function}
\begin{align*}
    h_{\Phi_n}(g_1,\mo_n(g_2),V^*_n)&=\vv{h_{\Phi_n,k_n}(g_1,\mo_n(g_2),V^*)} \\ &\vv{=\sum_{v\in V^*_n}\mathds{1}\left\{\text{rank}_{\Phi_n(g_1,\mo_n(g_2),V^*)}(\mo_n(v))\leq k_n\right\}}?
\end{align*}
Note that the scaling by $|V_n^*|$ in the recall setting and by $k_n$ in the precision setting do not affect consistent estimation of $h$ if $|V_n^*|=\Theta(1)$ or if in the precision setting $k_n=\Theta(1)$.  
 As such, the scaling is omitted.

The internal consistency criterion, Eq. \ref{eq:consis} guarantees that 
\begin{equation}
\label{eq:hmgh}
h_{\Phi_n}(g_1,\mo_n(g_2),V^*_n)=h_{\Phi_n}(g_1,\tilde{\mo}_n(g_2),V^*_n)
\end{equation}
for all obfuscation functions $\mo_n,\tilde{\mo}_n\in\mathfrak{O}_n$.
Indeed, the v.o.i.'s in $g_2$ are identical (though obfuscated differently) in $\mo_n(g_2)$ and $\tilde{\mo}_n(g_2)$.
If we consider an alternate $(g_1',g_2')\sim F'_n\subset \mathbf{F'}$, it could be the case that $g_1=g_1'$ and $g_2\simeq g_2'$, while 
\begin{equation}
    \label{eq:badh}
\vv{h_{\Phi_n}(g_1,\mo_n(g_2),V^*_n)\neq h_{\Phi_n}(g_1,\mo_n(g_2'),V^*_n)}
\end{equation}
\vv{for all $\mo_n\in\mathfrak{O}_n$}; indeed, consider letting the v.o.i.'s' in $g_2'$ be different from (and not isomorphic to) those in $g_2$ (i.e., the behavior of the v.o.i. in $F_n'$ is different from the behavior of the v.o.i. in $F_n$).

Consider the problem of estimating $h_{\Phi_n}$ via $\hat h_{\Phi_n}$.
If the estimator is label-agnostic (i.e., there is no information in the obfuscated labeling of $\mo(g_2)$), then it is sensible to require that \vv{for all $g_2\simeq g_2'$, we have that 
\begin{align}
\label{eq:hhat}
\hat h_{\Phi_n}(g_1,\mo_n(g_2),V^*_n)= \hat h_{\Phi_n}(g_1,\mo_n(g_2'),V^*_n).
\end{align}
Contrasting this to Eqs. (\ref{eq:hmgh}) and (\ref{eq:badh})}, we see that $(\hat h_{\Phi_n})$ cannot universally consistently estimate $(h_{\Phi_n})$, as the sequence of estimators cannot account for the potentially different behaviors of the v.o.i.'s under the umbrella of nominatable distributions.
To wit, we have the following lemma.
\begin{lemma}
\label{lem:verify}
With notation as above, let $(\hat h_{\Phi_n})_n$ be any sequence of label-agnostic (i.e., satisfying Eq.\@ \ref{eq:hhat}) estimators of $(h_{\Phi_n})_n$.
There exists sequences of nested-core nominatable distributions $\mathbf{F}=(F_n)$ and $\mathbf{F'}=(F_n')$ such that for $n$ sufficiently large,
if $(G_1,G_2)\sim F_n$, and $(G_1',G_2')\sim F_n'$, then 
$$
d_{\text{TV}}\left(\mathcal{L}(h_{\phi_n}(G_1,\mo(G_2),V^*_n)), \mathcal{L}(h_{\phi_n}(G_1',\mo(G_2'),V^*_n))\right)\textbf{}>0,$$
while $\mathcal{L}(\hat h_{\phi_n}(G_1,\mo(G_2),V^*_n))= \mathcal{L}(\hat h_{\phi_n}(G_1',\mo(G_2'),V^*_n))$ (where $d_{\text{TV}}$ is the total variation distance).  
\end{lemma}

As a result of the above discussion and Lemma, we are unable to verify, without additional supervision, if an adversary has moved the distribution out of a given VN rule's consistency class.
This points to the primacy of additional supervision, which in the VN framework often comes in the form of a user-in-the-loop.
Indeed, we are currently exploring the role/impact a use-in-the-loop in VN---where the user can evaluate the interestingness of the vertices in the top $k$ of the nomination list for a cost $c_k$.
This supervision can also be thought of as a form of regularization, designed to increase the consistency class of a given VN rule.

\section{Adversarial Vertex Nomination}
\label{sec:adv}
In order to actively model adversarial attacks in the VN-framework, we formalize the notion of an \textit{edge adversary}.
\begin{definition}[Adversary]
\label{def:Adv}
Let $F$ be a distribution on graphs in $\gm$, and let $U$ be a random variable independent of $G\sim F$.
 We say $\mathcal{A} = \{f_{\mathcal{A}}, V_{\mathcal{A}},U,\theta\}$ is an \textit{adversary} parameterized by $\theta\in\Theta$ if
\begin{enumerate}
\item{ $f_{\mathcal{A}}: \mathcal{G}_m \times \mathbb{R}\times \Theta\mapsto \mathcal{G}_m$ is a measurable function such that $V( f_{\mathcal{A}}(G,U,\theta)) = V(G),$
so that $f_{\mathcal{A}}(G,U,\theta)$ is a $\gm$-valued random variable.
}
\item{$V_{\mathcal{A}}:\mathcal{G}_m\times \mathbb{R}\times\Theta\mapsto 2^{[m]}$ is a measurable function that satisfies $V_{\mathcal{A}}(G,U,\theta) \subset V(G)$, so that
$V_{\mathcal{A}}(G,U,\theta)$ is a (potentially) random subset of $ V(G)$.}
\item{If $L = \bigg\{v,w \in V(G) \text{ s.t. }(v,w) \in E(f_{\mathcal{A}}(G,U,\theta))\, \Delta\, E(G) \bigg\},$ (where $\Delta$ represents the symmetric difference) then $L \subset V_{\mathcal{A}}(G,U,\theta)$.
Succinctly put, if an edge is added or removed from $E(G)$, then the vertices adjacent to that edge must be in $V_{\mathcal{A}}(G,U,\theta)$.}
\end{enumerate}
In the above, $U$ represents an independent source of randomness utilized in the adversarial attack.
\end{definition}
\noindent Note that $f_{\mathcal{A}}$ is simply a function that adds/deletes edges from a network potentially randomly, and these edges must be incident to the vertices of $V_{\mathcal{A}}$.  
To that end, we will refer to $V_{\mathcal{A}}$ as the vertices \textit{contaminated} by $\mathcal{A}$.   

If we are given a sequence of nominatable distributions ${\bf F}=(F_n)_{n=n_0}^{\infty}$, where $F_n$ is a distribution on $\mathcal{G}_n \times \mathcal{G}_m$, then we will let $f_{\mathcal{A}_n}(F_n)$ denote a sequence of graphs realized from $F_n$, with the second graph $G_2$ contaminated by $f_{\mathcal{A}_n}$; we call a sequence $(f_{\mathcal{A}_n})_{n=n_0}^{\infty}$ an adversary rule.
In the language of VN consistency classes, 
we posit that an adversary rule aims to contaminate a VN rule $\mathbf{\Phi}$ via
$$
{\bf F}=(F_n)_{n=n_0}^\infty\in \mathfrak{C}_{{\mathbf{\Phi}}}^{(k_n)}\implies 
(f_{\mathcal{A}_n}(F_n))_{n=n_0}^\infty\in\mathfrak{N}_{\bf V^*}\setminus \mathfrak{C}_{\mathbf{\Phi}}^{(k_n)}.
$$

\begin{remark}
Let $G_2=(V_2,E_2)$ and $G_2'=(V_2',E_2')$.
Consider an edge adversary $\fa$ acting on $G_2'$.
By considering $V_2=V(G_2')\setminus \va$, we can also consider this adversary as a \emph{vertex adversary} that randomly adds vertices to $G_2$.
Vertex addition and deletion can be simultaneously modeled by first considering a mechanism for randomly deleting vertices from $G_2=(V_2,E_2)$ before using the above approach to add adversarial vertices to the network.
\end{remark}

\begin{remark}
In \cite{adv2}, the authors consider \textit{direct attacks} and \textit{influencer attacks} in which, given a vertex of interest $v^*$, either $v^* \in V_{\mathcal{A}}$ or $v^* \notin V_{\mathcal{A}}$ respectively.  However, note that in \cite{adv2}, the objective is vertex classification, whereas we are not directly classifying vertices.
Rather, we are interested in ranking vertices in $G_2$ by interestingness given limited training data in $G_1$.  
We will typically assume that $v^* \notin V_{\mathcal{A}}$ (i.e. the adversary does not control the vertex of interest), so that we are examining \textit{influencer attacks}.
\end{remark}

\subsection{A Simple VN Adversarial Contamination Model}
\label{sec:model}
Now that we have developed the requisite \vv{setting} for framing the idea of adversarial contamination in the VN-setting, we will consider a simple model for adversarial contamination in the stochastic blockmodel (SBM) of \cite{sbm}.  
\begin{definition}[Stochastic Blockmodel]
We say that an $n$-vertex random graph $G$ is an instantiation of a stochastic blockmodel with parameters $(n,K,B,b)$ (written $A\sim\text{SBM}(n,K,B,\pi)$) if 
\begin{itemize}
\item[i.] The block membership vector $\pi\in\mathbb{R}^K$ satisfies $\pi_i\geq 0$ for all $i\in[K]$, and $\sum_i\pi(i)=~1$;
\item[ii.] The vertex set $V=V(G)$ is the disjoint union of $K$ blocks
$V=B_1 \sqcup B_2   \sqcup \cdots \sqcup B_K$, where each vertex $v\in V$ is independently assigned to a block according to a Multinomial($1,\pi$) distribution.
If vertex $v$ is assigned to block $i\in[K]$, then the block membership function $b:V\mapsto[K]$ satisfies $b(v)=i$;
\item[iii.] The block probability matrix $B\in[0,1]^{K\times K}$ is such that, for each pair
of vertices $\{u,v\}\in\binom{V}{2}$, $\mathds{1}_{u\sim_{G} v}\sim\text{Bernoulli}(B_{b(u),b(v)})$, and the collection of indicator random variables
$\{\mathds{1}_{u\sim_{G} v}\}_{\{u,v\}\in\binom{V}{2}}$
is mutually independent \vv{given $b$} (here $\{u\sim_{G} v\}\Leftrightarrow\{\{u,v\}\in E\}$).
\end{itemize}
In addition, we will say that a pair of graphs $(G_1,G_2)$ is an instantiation of a $\rho$-correlated $\text{SBM}(n,K,B,b)$ (written $(G_1,G_2)\sim \text{SBM}(\rho,n,K,B,\pi)$) if marginally $G_1\sim\text{SBM}(n,K,B,b)$ and $G_2\sim\text{SBM}(n,K,B,b)$, and the collection of indicator random variables
$$\left\{\{\mathds{1}_{u\sim_{G_1} v}\}_{\{u,v\}\in\binom{V}{2}}\bigcup \{\mathds{1}_{u\sim_{G_2} v}\}_{\{u,v\}\in\binom{V}{2}}
\right\}$$
is mutually independent except that for each $\{u,v\}\in\binom{V}{2}$, $$\text{Correlation}(\mathds{1}_{u\sim_{G_1} v},\mathds{1}_{u\sim_{G_2} v})=\rho.$$
\end{definition}

Consider $G$ as an $n$-vertex stochastic blockmodel, with two blocks, $B_1$ and $B_2$, and with $\pi=(1/2,1/2)$ .  The block-probability matrix $B$ is given by 
\begin{align}
\label{eq:B}
B = \begin{pmatrix} p & r \\ r & q  \end{pmatrix},
\end{align} with $p \geq q \geq r > 0$.
Given $G=g$, we define the following VN adversarial contamination procedure $\mathcal{A}=(\fa,V_\mathcal{A},U,\theta)$ acting on $g$ as follows:
\begin{enumerate}
\item{$\theta=(c_+, c_-,\pi_+,\pi_-,s_+,s_-)$ is a vector of parameters where $c_+, c_-\in\mathbb{Z}$ satisfy $c_++c_-\leq n$, $\pi_+,\,\pi_-\in(0,1)$, and $s_+$, $s_-\in[0,1]$;}

\item{ $U$ is a uniformly distributed random variable independent of $G$;}

\item{$\fa(g,U,\theta)\in\mathcal{G}_n$ is defined as follows:}
\begin{enumerate}
\item[i.]{Initialize $g_c=g$}
\item[ii.]\vv{Create a set of vertices $W_+$ by independently selecting each vertex in $V=[n]$ to be in $W_+$ with probability $\pi_+$.
Then, create a set of vertices $W_-$ by independently selecting each vertex in $V\setminus W_+=[n]$ to be in $W_-$ with probability $\pi_-$.}

\item[iii.]{ For each vertex pair $\{v,u\} \in W_+\times (V \setminus W_-)$,}
\begin{enumerate}
\item[i.]{If $\{v,u\}\in E(g_c)$, nothing happens.}
\item[ii.]{If $\{v,u\}\notin E(g_c)$, an edge is independently added connecting $\{v,u\}$ in $g_c$ with probability $s_+$.}
\end{enumerate}
\item[iv.]{For each vertex pair $\{v,u\} \in W_-\times (V \setminus W_+)$,}
\begin{enumerate}
\item{If $\{v,u\}\notin E(g_c)$, nothing happens.}
\item{If $\{v,u\}\in E(g_c)$, the edge is independently deleted from $g_c$ with probability $s_-$.}
\end{enumerate}
\item[v.] Set $\fa(g,U,\theta)=g_c\in\mathcal{G}_n.$
\end{enumerate}
\end{enumerate}
The auxiliary randomness $U$ in $\mathcal{A}$ is utilized to make the random vertex selections in ii., the random edge additions in iii., and the random edge deletions in iv.

Notice that this adversarial model gives rise to a new stochastic blockmodel with the edge-probability matrix $\tilde B$ given by 
\[\tilde B =
\begin{blockarray}{ccccccc}
& \tilde  B_1 &  \tilde B_1^+ &  \tilde B_1^- &  \tilde B_2 &  \tilde B_2^+ &  \tilde B_2^- \\
\begin{block}{c(cccccc)}
\tilde B_1       & {\bf p}  & x_1  & x_2  & {\bf r}   &  x_3  & x_4    \\
\tilde B_1^+     &x_1   &x_1  & p  & x_3   & x_5  & r   \\
\tilde B_1^-     & x_2  & p  & x_2& x_4  &r    & x_6\\
 \tilde B_2      & {\bf r}  & x_3  & x_4 & {\bf q}   & x_{7}&  x_{8}  \\
 \tilde B_2^+    &  x_3 & x_5 &r   & x_{7}& x_{7}& q    \\
 \tilde B_2^-    & x_4 & r  & x_6& x_{8}& q   & x_{8} \\
\end{block}
\end{blockarray}
 \]
 where
 \begin{align*}
 x_1&= p + s_+(1-p),\hspace{2mm} x_2=p(1-s_{-}),\hspace{2mm} x_3=r + s_+(1-r),\\
x_4&=(1-s_-)r,\hspace{2mm}x_5=r+(2s_+-s_+)^2(1-r),\\ x_6 &=r(1-s_-)^2, 
x_{7}=q+s_+(1-q),\hspace{2mm} x_{8}=q(1-s_-),
\end{align*}
and where $ \tilde B_1^+$ are the vertices in $W_+ \cap B_1$; $ \tilde B_1^-$ are the vertices in $B_1 \cap W_-$; and $ \tilde B_1$ are the vertices in $B_1\setminus ( \tilde B_1^+ \cup  \tilde B_1^-)$; with $ \tilde B_2$ defined analogously.  
We note here that this adversarial contamination model is similar to the contamination model considered in \cite{cai2015robust}.

Note also that the original block structure is preserved amongst vertices in $\tilde B_1\cup \tilde B_2$, and we can view this contamination model as adding vertices randomly to $G[\tilde B_1\cup \tilde B_2]$, i.e., the induced subgraph on $\tilde B_1\cup \tilde B_2$.
When $(G_1,G_2)\sim \text{SBM}(\rho,n,K,B,\pi)$ and this adversarial procedure is applied to $G_2$, we will denote 
\begin{align}
\label{eq:gs1}
G_1^{(i)}&=G_1[\tilde B_1\cup \tilde B_2]\\
\label{eq:gs2}
G_2^{(i)}&=G_2[\tilde B_1\cup \tilde B_2]
\end{align}

\begin{remark}
\label{rem:inconsistent}
Let $\mathcal{A}_n$ be the simple adversarial rule outlined above.
A very simple VN rule $\mathbf{\Phi}$ and nested core nominatable sequence ${\bf F}$ for which 
$$
{\bf F}=(F_n)_{n=n_0}^\infty\in \mathfrak{C}_{{\mathbf{\Phi}}}^{(k_n)}\implies 
(f_{\mathcal{A}_n}(F_n))_{n=n_0}^\infty\in\mathfrak{N}\setminus \mathfrak{C}_{\mathbf{\Phi}}^{(k_n)}.
$$ 
proceeds as follows.
Consider $F_n=\text{SBM}(\rho,n,K,B,\pi)$ supported on $\gn\times\gn$ where $B$ is as in Eq. \ref{eq:B} with $\pi=(1/2,1/2)$, $p>q>r$ fixed, and $\rho>0$ fixed.
Suppose that $\Phi_n$ is a VN scheme that runs spectral clustering on the contaminated graph by first selecting the number of communities in a consistent manner (via adjacency spectral clustering for example \cite{perfect}) and ranking all the vertices in the group with the highest probability of within-group connection (in a fixed but arbitrary order), and then ranks the rest of the vertices in fixed but arbitrary order.
Suppose that we consider $k_n = n/2$.  
It is immediate that ${\bf F}=(F_n)_{n=n_0}^\infty\in \mathfrak{C}_{{\mathbf{\Phi}}}^{(k_n)}$ and that
the adversary acting on $G_2$ impacts this consistency. We present the following result as a lemma, but the proof is a simple calculation.
\end{remark}
\begin{lemma}In the adversarial contamination model $\mathcal{A}_n$ defined above, 
if either
\begin{enumerate}
\item $p-q < s_-$, or
\item $\frac{p-q}{1-q} < s_+$,
\end{enumerate}
then $\Phi_n$ is no longer consistent with respect to the adversarially contaminated model sequence.
\end{lemma}

\subsection{Regularizing the Adversary}
\label{sec:reg}
Given the adversarial model considered above, and the discussion on VN verification in Section \ref{sec:verify}, it is natural to seek procedures for mitigating the effect of the contamination in $G_2$.  
Network regularization is a natural solution, and we here consider as a regularization strategy the network analogue of the classical trimmed mean estimator.
To wit, we consider the regularization procedure in Algorithm \ref{alg:trim} inspired by the network trimming procedure in \cite{edge2018trimming}; 
see also the work in \cite{le2017concentration} for the impact of trimming regularization on random graph concentration.
\begin{algorithm}[h!]
  \begin{algorithmic}
    \STATE \textbf{Input}: Graph $G$, $\ell,h\in(0,1)$, seed set $S$;
\vspace{3mm}
\STATE{\bf 1.} Initialize $V_t=S$
\STATE{\bf 2.} Rank the vertices in $V(G)\setminus S$ by descending degree (ties are broken via averaging over ranks).
For each vertex $u$ in $V(G)\setminus S$, denote the rank via $rk(u)$;
\FOR{$u\in V(G)\setminus S$,}
\STATE{\bf 3.} If $\ell<\frac{rk(u)}{|V(G)\setminus S|}\leq 1-h$, add $u$ to $V_t;$
\ENDFOR
\STATE{\bf 4.} \textbf{Output}: $G^{(\ell,h)}=G[V_t]$, the induced subgraph of $G$ on $V_t$;
\end{algorithmic}
\caption{Regularization via network trimming}
\label{alg:trim}
\end{algorithm}
\begin{remark}
\label{rem:modularity}
The parameters $\ell$ and $h$ appearing in Algorithm \ref{alg:trim} are unknown a priori, and to data-adaptively choose $\ell$ and $h$, we sweep over possible values and choose the values of $\ell$ and $h$ that leads to the maximum network modularity in $G_2^{(\ell,h)}$ when clustering the vertices of $G_2^{(\ell,h)}$ via  $GMM\circ ASE$ clustering; i.e., embed $G_2^{(\ell,h)}$ using ASE and cluster the embedding using a model-based GMM procedure.
Given a clustering $C$, the modularity is defined as usual via
$$
Q(C)={\frac {1}{(2|E|)}}\sum _{i,j}\left[A_{i,j}-{\frac {d_{i}d_{j}}{2|E|}}\right]\mathds{1}\{C_{i}=C_{j}\},
$$
where
$|E|=$the number of edges in $G_2^{(\ell,h)}$; $A_{i,j}$ is the $i,j$-th element of the adjacency matrix $A$ of $G_2^{(\ell,h)}$; $d_i$ is the degree of vertex $i$ in $G_2^{(\ell,h)}$; and $C_i$ is the cluster containing vertex $i$ in $C$.
\end{remark}

\subsubsection{Regularization in our Motivating Example from Section \ref{sec:Motive}}
\label{sec:regreg}
\begin{figure*}[t!]
\centering
\subfloat{\includegraphics[width=0.47\textwidth]{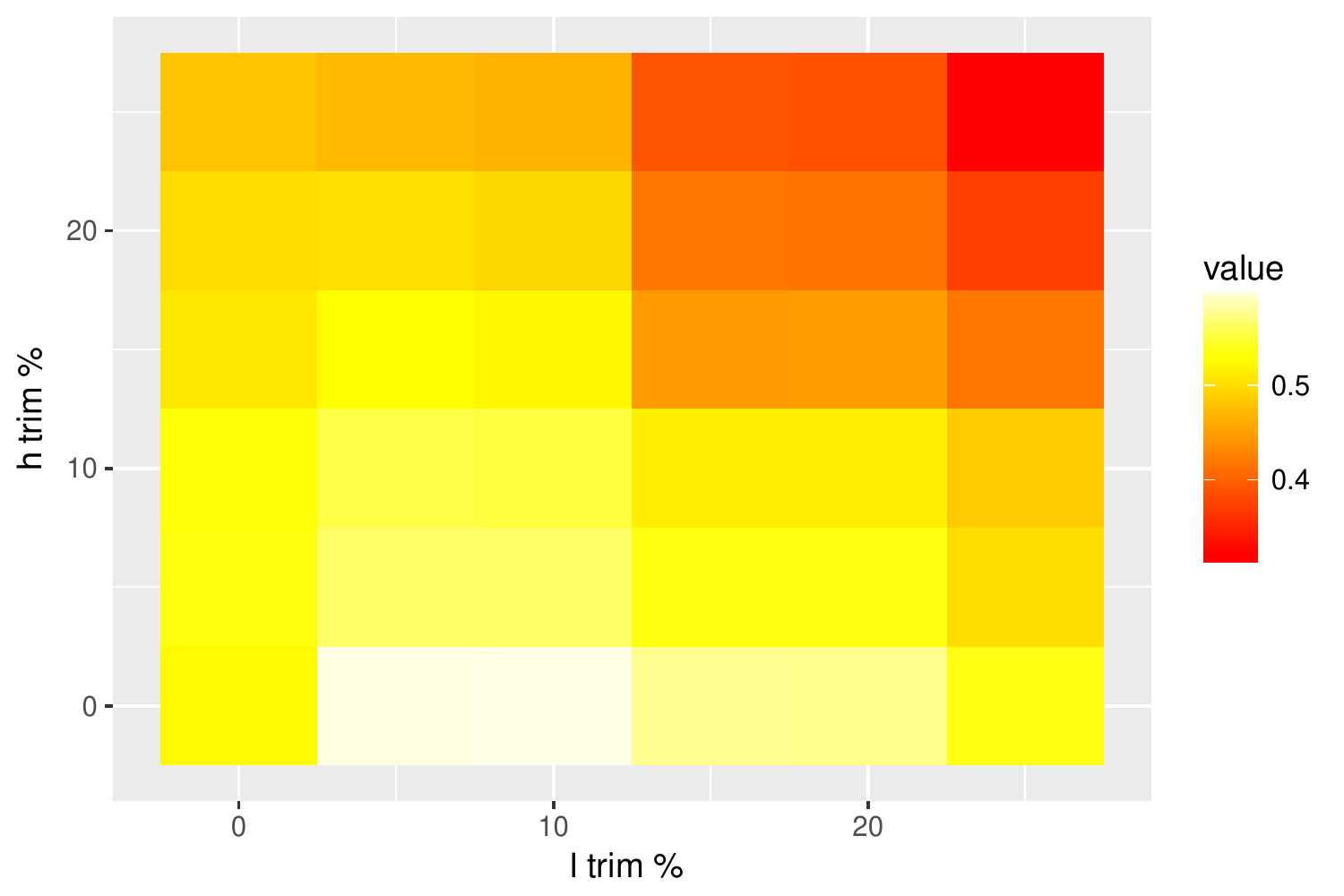}}
\hfil
\subfloat{	\includegraphics[width=0.47\textwidth]{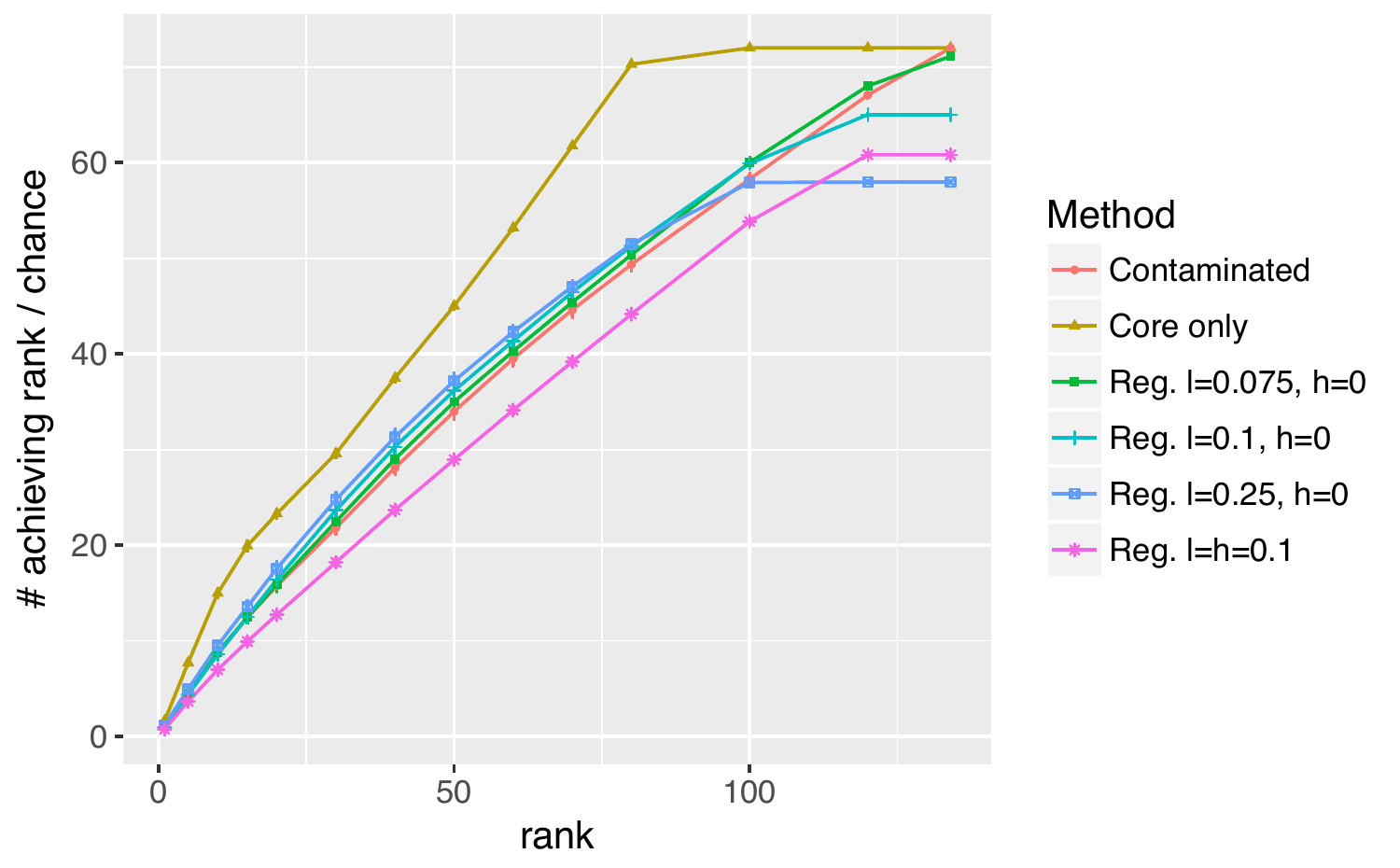}}
\caption{The left panel shows the modularity of the GMM clustering as a function of the regularization parameters.  
The color indicates the value of the modularity, with darker red indicating lower values and lighter yellow--to--white indicating larger values.
The right panel shows the performance of $\VNA$ with higher values indicating a greater percentage of true vertices nominated.
In both figures, we average over the same 500 seed sets of size $s=10$.  See Section \ref{sec:regreg} for details.}
	\label{fig:reg}
\end{figure*}

We next explore the impact of regularization on our motivating HS social network example from Section \ref{sec:Motive}.
In the left panel of Figure \ref{fig:reg}, we plot the modularity of the GMM clustering in the trimmed $G_2^{(\ell,h)}$ as a function of $\ell,h\in\{0,0.05,0.1,0.15,0.2,0.25\}$.
Note that we average the modularity values over $nMC=500$ seed sets of size $s=10$ (the same seed sets as used in Figure \ref{fig:vnfex}).
The color indicates the value of the modularity, with darker red indicating lower values and lighter yellow--to--white indicating larger values.
From the figure, we can see that modularity is maximized when $h=0$ (i.e., no large degree vertices trimmed) and $\ell\approx 0.05$--$0.1$.
We note that this trimming process can cut core vertices as well as junk vertices, and 
core vertices cut from $G_2$ can never be recovered via $\VNA$. 
This is demonstrated in the right panel of Figure \ref{fig:reg}, where the horizontal asymptotes for each trimming value indicates the maximum number of core vertices that are recoverable after regularization. 
In the figure, the gold line represents performance in the idealized network pair $(G_1^{(i)},G_2^{(i)})$; the red line for the contaminated $(G_1^{(i)},G_2)$;
the green, teal, blue and pink lines respectively present performance for $(G_1,G_2^{(\ell,h)})$ (i.e., after regularizing) with $(\ell,h)=(0.075,0),$ $(0.1,0),$ $(0.25,0),$ $(0.1,0.1)$ respectively.

\begin{table}
\begin{tabular}{c|c|c|c|c|c|c}
&\multicolumn{6}{c}{Mean number of v.o.i. achieving rank $\leq$ x}\\
& $x=1$ & $x=5$ & $x=10$ & $x=15$ & $x=20$ & $x=30$\\
\hline
Core only&2.800 &12.624 &21.882& 27.448& 31.882& 38.884 \\
Contaminated &2.074 &10.006& 17.346& 21.428& 24.656& 29.598 \\
Reg. $\ell=0.075, h=0$& 1.920&  9.500 &16.198& 20.536& 24.372& 31.266\\ 
Reg. $\ell=0.1, h=0$& 1.580&  8.296 &14.216& 19.918& 25.004& 34.434\\ 
Reg. $\ell=0.25, h=0$& 1.572&  8.274 &15.146& 21.136& 26.792& 36.630\\ 
Reg. $\ell=0.1, h=0.1$& 1.970&  8.756 &13.678& 17.284& 20.470& 26.574
\end{tabular}
\caption{Mean number of v.o.i. achieving rank $\leq$ x for the various regularization and contamination settings considered.  Results are averaged over 500 MC trials.}
\label{table:HSfriend}
\end{table}

\begin{figure*}[!t]
\centering
\subfloat{\includegraphics[width=0.45\textwidth]{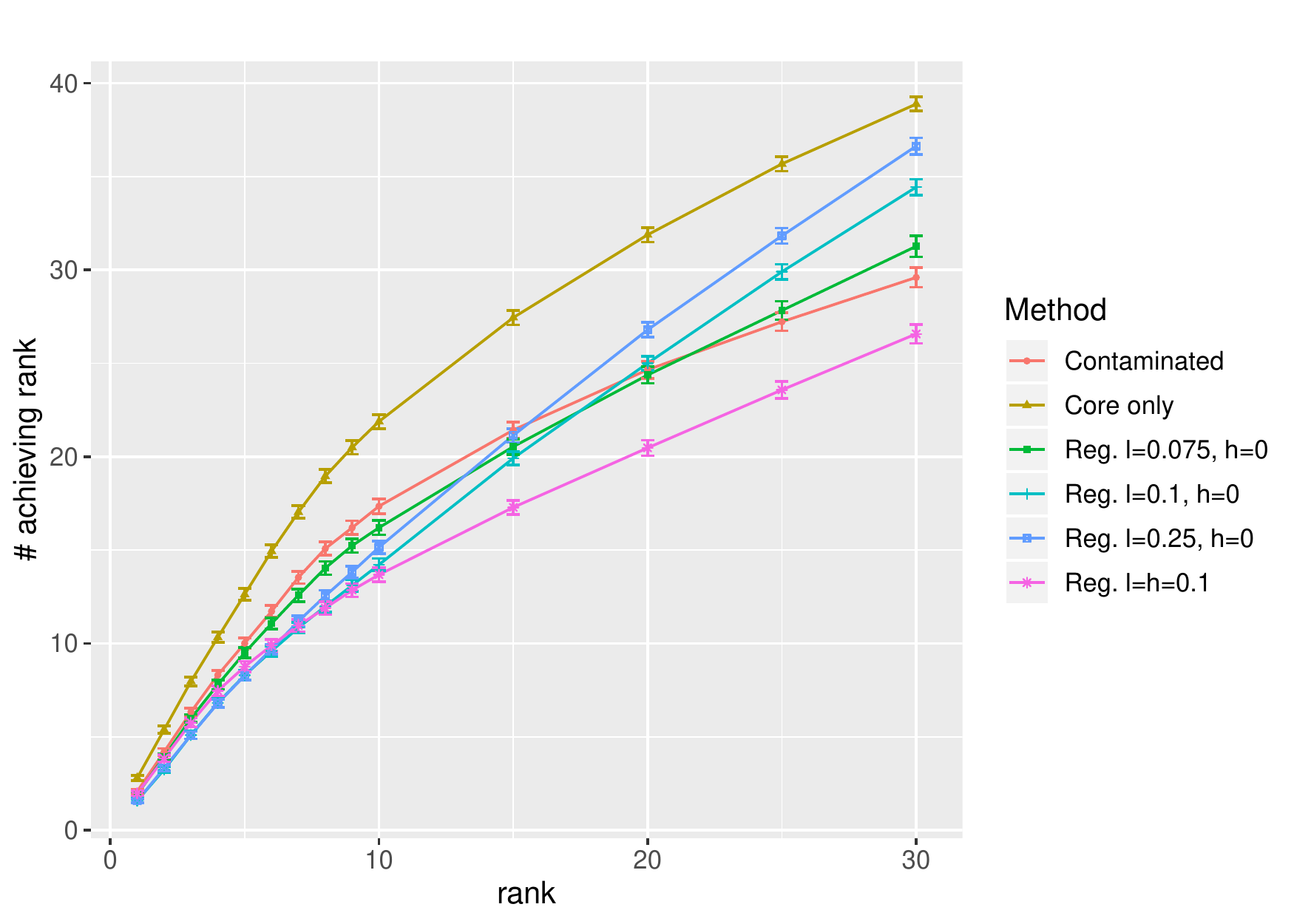}
\label{fig_first_case}}
\hfil
\subfloat{\includegraphics[width=0.45\textwidth]{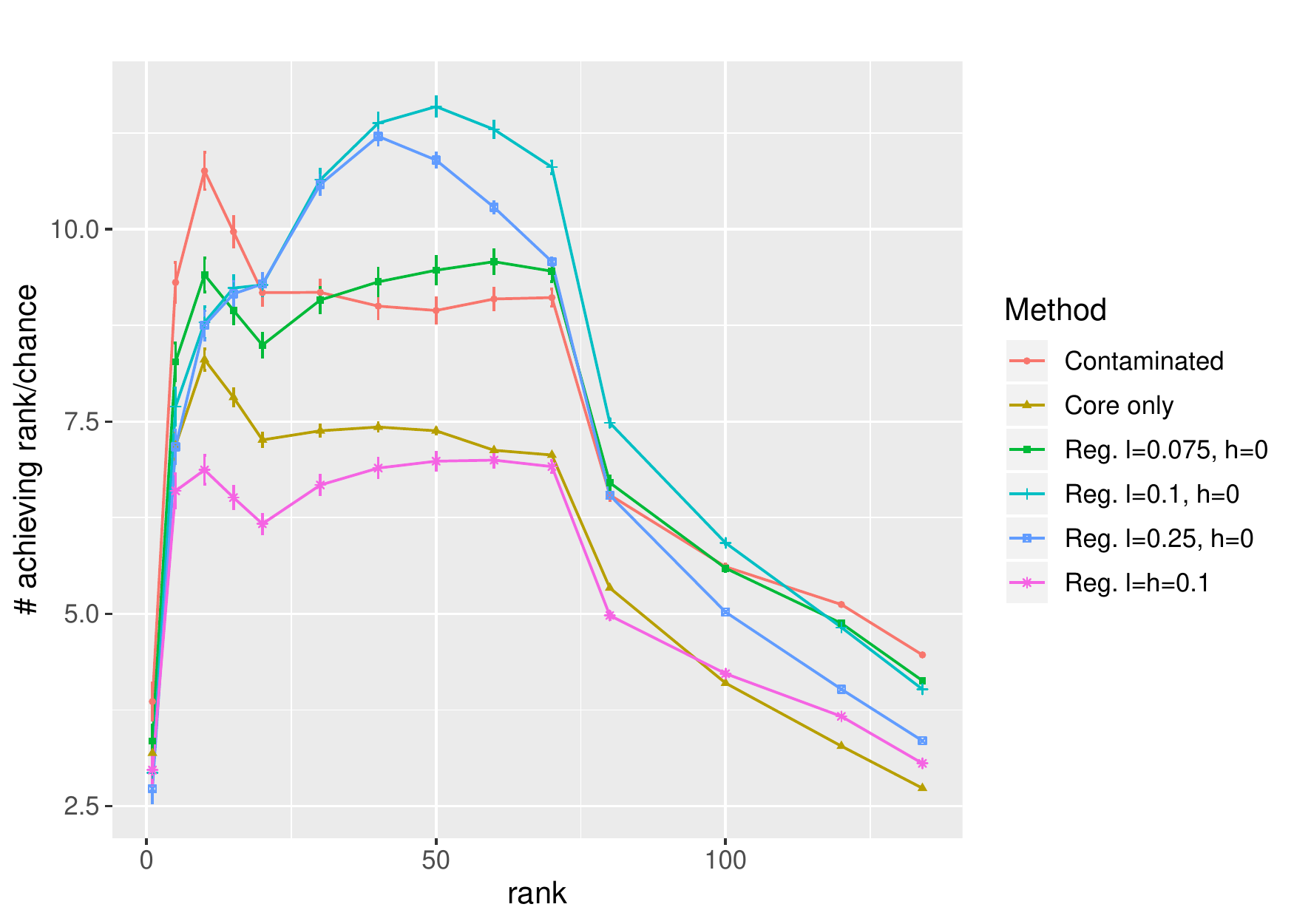}
\label{fig_second_case}}
\caption{
We plot the performance of $\VNA$ averaged over $nMC=500$ random seed sets of size $s=10$.  
The left figure shows the number of true vertices achieving the rank, and the right figure shows the same result normalized by chance performance.  
The gold line represents the original, uncontaminated netowrk, the red represents the contaminated network, and the other colors represent differerent levels of regularization.  See Section \ref{sec:regreg} for details.
}
\label{fig:vnfex2}
\end{figure*}

In Figure \ref{fig:vnfex2} and Table \ref{table:HSfriend}, we see the effect of regularization play out in more detail.
Indeed, mean $\VNA$ performance in the regularized setting increases versus in the contaminated setting for $$(\ell,h)=\{(0.075,0),(0.1,0),(0.25,0)\},$$ whereas mean regularized performance decreases for $(\ell,h)=\{(0.1,0.1)\}$.
While over-regularizing can adversely affect performance, this data-adaptive regularization --- while not fully recovering the performance of the idealized setting---nonetheless effectively mitigates the impact of the contamination on our $\VNA$ algorithm in this dataset.

\section{Experiments}
\label{sec:data}
We next explore the effect of our adversarial noise model in a simulated data experiment, and the effect of adversarial contamination (and a subsequent model for regularization) in a real data example derived from Bing entity transition graphs. 
First, we explain in detail the steps of the VN scheme we will consider in our experiments.

\subsection{Experimental Setup}
\label{sec:asegmm}

In the contamination model of Section \ref{sec:model}, we consider the following VN scheme, denoted $\VNA$.
Letting $v^*\in V(G_1)$ (resp., $V^*\subset V(G_1)$) be the vertex (resp., vertices) of interest in $G_1$, we seek the corresponding vertex (resp., vertices) of interest in $V(G_2)$ as follows:

\noindent{\bf 1.} Given two graphs, $G_1$ and $G_2$, we use Adjacency Spectral Embedding (ASE) \cite{sussman2014consistent} to separately embed $G_1$ and $G_2$ into a common Euclidean space $\mathbb{R}^d$.
Given the $n\times n$ adjacency matrix $A$ of $G_1$, the $d$-dimensional ASE of $G_1$ is defined as follows.
\begin{definition} [Adjacency spectral embedding (ASE)]\label{def:ASE}
	Given $d \in\mathbb{Z}>0$, the {\em adjacency spectral embedding} (ASE) of $A$ into $\mathbb{R}^{d}$ is defined via $\widehat{{ X}}={ U}_{{A}}
	{ S}_{{A}}^{1/2}$ where
	$$|{ A}|=[{ U}_{{A}}|{ U}^{\perp}_{{A}}][{
		S}_{{A}} \oplus { S}^{\perp}_{{A}}][{
		U}_{{A}}|{ U}^{\perp}_{{A}}]$$ is the spectral
	decomposition of $|{A}| = ({A}^{T}{A})^{1/2}$, 
	${S}_{{A}}\in\mathbb{R}^{d\times d}$ is the diagonal matrix with the $d$ largest eigenvalues
	of $|{A}|$ on its diagonal and ${U}_{{A}}\in\mathbb{R}^{n\times d}$  has 
	columns which are the eigenvectors corresponding to the eigenvalues of ${S}_{{A}}$.
\end{definition}
\noindent Simply stated, the ASE of a graph $G$ provides Euclidean features for each vertex in $G$ on which to perform subsequent inference.
Combined with recent efforts to prove that the ASE provides consistent estimators of the latent position parameters in random dot product graphs and positive-definite stochastic blockmodels \cite{sussman2014consistent,rdpgsurvey}, the ASE allows for a host classical inference methodologies to be successfully employed within these random graph frameworks \cite{MT2,tang2014nonparametric,lyzinski2015community}.
To choose $d$ above, we use the machinery of \cite{zhu2006automatic,chatterjee2014matrix} to develop the principled heuristic of estimating $d$ as the larger of the two elbows of the associated scree plots of the singular values of $G_1$ and $G_2$.
\vspace{3mm}

\noindent {\bf 2.} Solve the orthogonal Procrustes problem \cite{schonemann1966generalized} to find an orthogonal transformation aligning the seeded vertices across graphs.
Let $\widehat{X}_S$ (resp., $\widehat{Y}_S$) be the matrix composed of the rows of ASE($G_1$) (resp., ASE($G_2$)) corresponding to the seeded vertices in $S$.
Letting the SVD of $\widehat Y_S^T\widehat X_S=U\Sigma V^T$, the solution to 
$$R=\text{argmin} _{O\text{ s.t. }O ^{T}O =I }\| \widehat X_S-\widehat Y_S O\|_{F},$$
is given by $R=UV^T$.
Use this transformation to align the embeddings of $G_1$ and $G_2$ in $\mathbb{R}^d$, i.e., rotate $\widehat Y$ via $\widehat Y O $ to align $\widehat Y$ to $\widehat X$.
\vspace{3mm}

\noindent {\bf 3.} Motivated by the central limit theorem of \cite{clt} for the residual errors between the rows of the ASE and the latent position parameters in random dot product graphs, we use model-based Gaussian mixture modeling (GMM) to simultaneously cluster the vertices of the embedded graphs.
Here, we employ the \texttt{R} package \texttt{MClust} \cite{fraley1999mclust}.

\vspace{3mm}

\noindent {\bf 4.} Rank the candidate matches in $G_2$ according to the following heuristic.
If $u\in V(G_1)$ and $v\in V(G_2)$ are clustered points in the Procrustes-aligned embedding of $G_1$ and $G_2$ with respective covariance matrices $\Sigma_u$ and $\Sigma_v$ in their components of the GMM, then compute $$\Delta(u,v)=\max \left(D_u(u,v),D_v(u,v)\right),$$
where $$D_u(u,v)=\sqrt{(u-v)\Sigma_u^{-1}(u-v)^T }$$ and $$D_v(u,v)=\sqrt{(u-v)\Sigma_v^{-1}(u-v)^T }$$
are the respective Mahalanobis distances from $u$ to $v$.
In the case of a single v.o.i. $v^*$, rank the vertices in $G_2$ then by increasing value of $\Delta(v^*,u)$, i.e., with ties broken in a fixed deterministic fashion, we rank via (where $n_2=|V(G_2)|$)
\begin{align*}
\Phi_n(g_1,g_2,v^*)[1]&\in\argmin_{u\in V(G_2)}\Delta(v^*,u)\\
\Phi_n(g_1,g_2,v^*)[2]&\in\argmin_{u\in V(G_2)\setminus\{\Phi_n[1] \}}\Delta(v^*,u)\\
&\hspace{10mm}\vdots\\
\Phi_n(g_1,g_2,v^*)[n_2-1]&\in\argmin_{u\in V(G_2) \setminus\{\cup_{(j\leq n_2-2)} \Phi_n[j] \}}\Delta(v^*,u)\\
\Phi_n(g_1,g_2,v^*)[n_2]&\in\argmin_{u\in C_{v^*}\setminus\{\cup_{(j\leq n_2-1)}\Phi_n[j]\}}\Delta(v^*,u).
\end{align*}
In the case of multiple v.o.i. $V^*$, rank the vertices in $G_2$ then by increasing value of $\min_{v\in V^*}\Delta(v,u)$ with ties broken in a fixed deterministic fashion.
\vv{We choose $\min_{v\in V^*}\Delta(v,u)$ as our ranking metric here as what defines interestingness can vary even among the v.o.i. in $G_1$; i.e., $\max_{v,v'\in V^*}\Delta(v,v')$ may be relatively large.  
Being uniformly close to the collection of v.o.i. would be too stringent a condition then, and we merely require highly nominated vertices to have close proximity to a v.o.i., as this would be evidence the highly nominated vertices correspond in $G_2$ to these proximal v.o.i. in $G_1$.}
\vspace{2mm}

\begin{figure*}[t]
	\centering
\subfloat[Modularity plot versus $(\ell,h)$]{\includegraphics[width=0.45\textwidth]{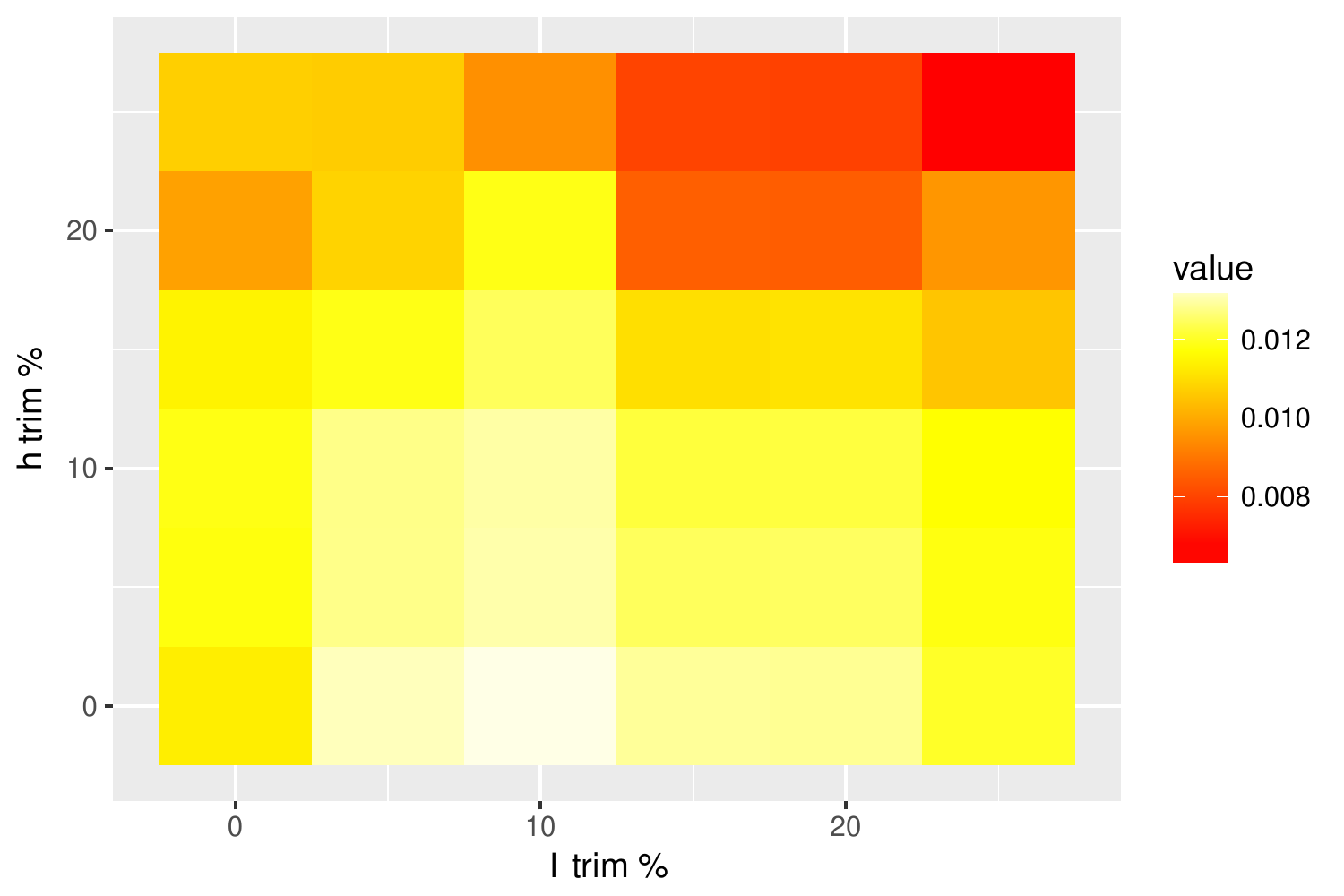}}
\subfloat[Regularization in $\VNA$]{\includegraphics[width=0.5\textwidth]{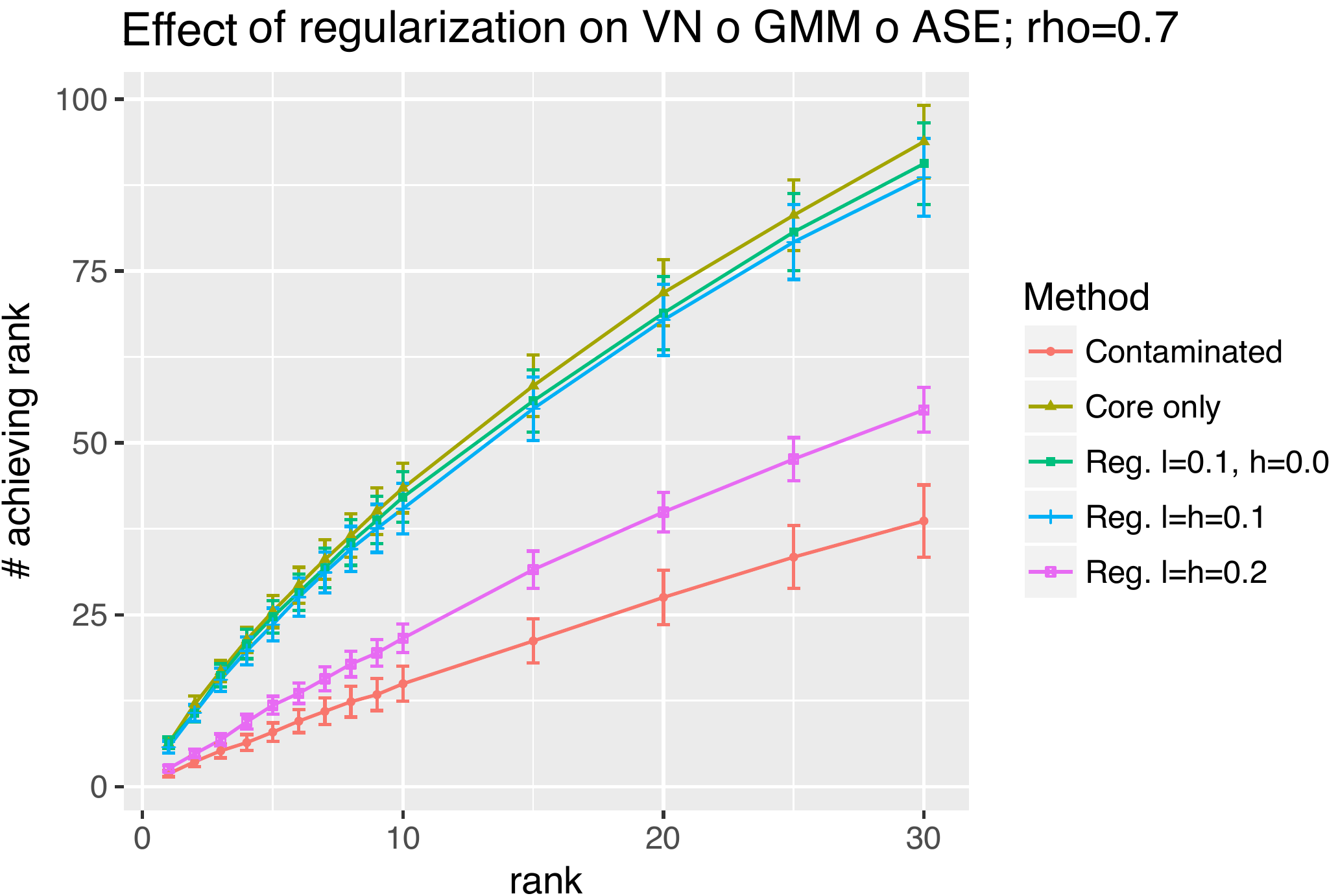} }
	
	\caption{
In the left panel, we plot the modularity of the GMM clustering in the trimmed graph as a function of the regularization parameters. 
The color indicates the value of the modularity, with darker red indicating lower values and lighter yellow--white indicating larger values.
In the right panel, we plot the performance of $\VNA$ for a stochastic blockmodel in our contamination model in Section \ref{sec:model}.  Gold represents the uncontaminated network, red represents the contaminated network, and the other colors represent various levels of reuglarization.  
In both figures, we average over the same 50 seed sets of size $s=10$. See Section \ref{sec: synth} for details.}
	\label{fig:simsim}
\end{figure*}

\begin{figure*}[t]
	\centering
\includegraphics[width=0.51\textwidth]{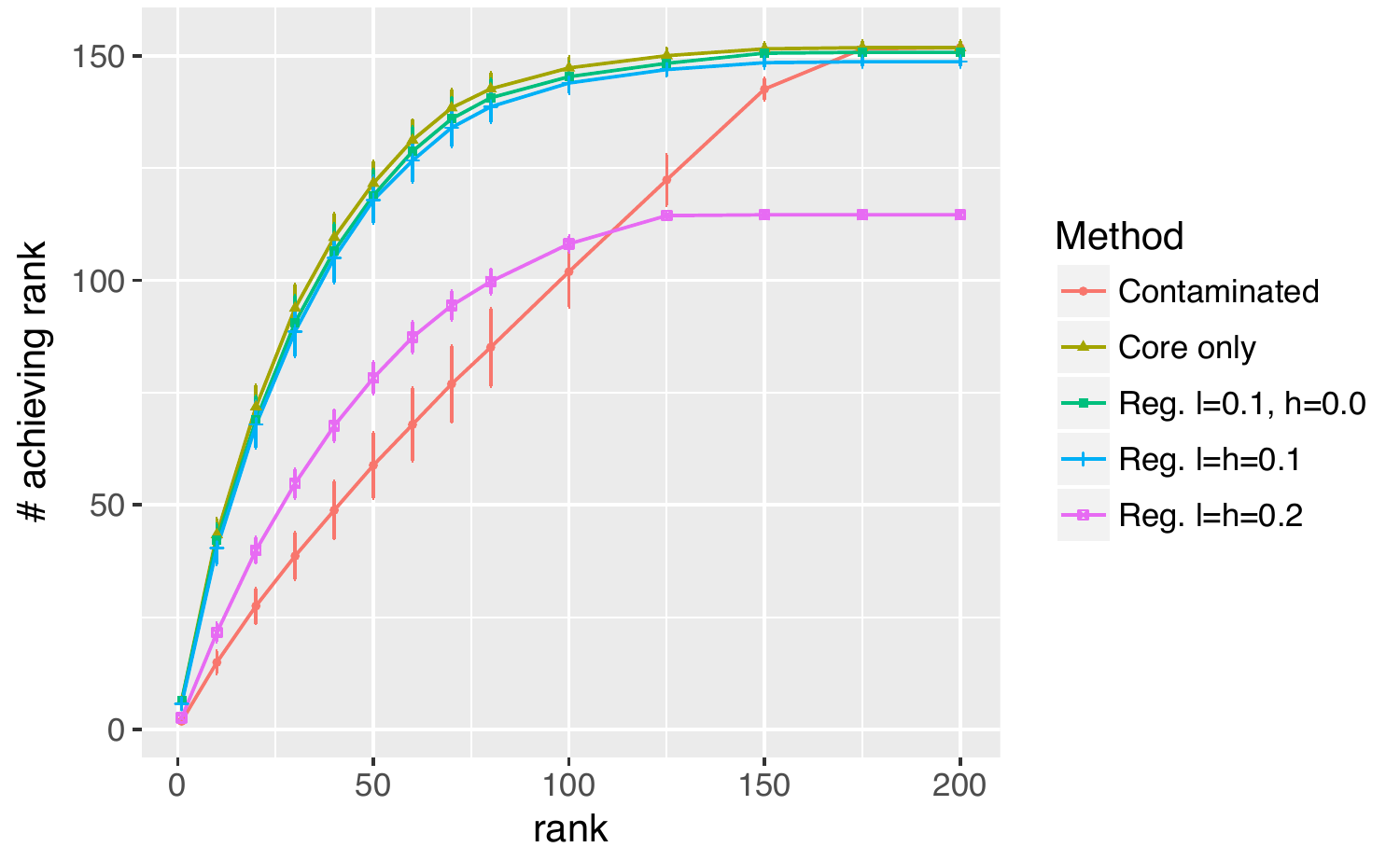}
	\caption{
We plot
the performance of $\VNA$ ($\pm$2s.e.) in $(G_1,G_2)\sim\text{SBM}(0.3,200,2,B,\pi=(1/2,1/2))$ with $\rho=0.7$ again averaged over $nMC=50$ random seed sets of size $s=10$.
The $x$-axis shows the ranks in the nomination list and the $y$-axis shows (on average) how many vertices $v\in G_1^{(i)}$, when viewed as the v.o.i., had their corresponding vertex of interest ranked in the top $x$ by $\VNA$. 
The gold line represents performance in the idealized network pair $(G_1^{(i)},G_2^{(i)})$; the red line for $(G_1^{(i)},G_2^{(c)})$; 
the green, blue, and pink lines for $(G_1^{(i)},G_2^{(\ell,h)})$
for varying values of $(\ell,h)$.}
	\label{fig:simtrim}
\end{figure*}

\subsection{Simulation}
\label{sec: synth}

We consider the model in Section \ref{sec:model} with the following parameter choices:
\[
\begin{tabular}{ccc}
$n=200$;& $\pi=(1/2,1/2)$;& $\pi_-=0.1,\,\, \pi_+=0.1$;\\
$p=0.4$;&$q=0.5$;&$r=0.3$;\\
$s_+=0.8$;& $s_-=0.8$;& $\rho\in(0.3, 0.5, 0.7).$
\end{tabular}
\]
\vv{Note that these parameter choices yield an illustrative simulation, and we find that the resulting findings hold across multiple parameter choices as well.}
Note that, in the notation of Section \ref{sec:model}, if $(G_1,G_2)\sim\text{SBM}(\rho,n,K,B,\pi)$, we will consider
\begin{align*}
G_1^{(i)}&=G_1[\tilde B_1\cup \tilde B_2]\\
G_2^{(i)}&=G_2[\tilde B_1\cup \tilde B_2]\\
G_2^{(c)}&=G_2\text{ acted upon by the adversary described} \\&\hspace{40mm}\text{ in Section }\ref{sec:model};\\
G_2^{(\ell,h)}&=G_2^{(c)}\text{ trimmed as in Algorithm }\ref{alg:trim}.
\end{align*}
In this simulation example, we observe that the adversarial contamination model significantly decreases VN performance and that the trimming regularization mitigates this contamination and recovers much of the lost inferential performance.

In Figure \ref{fig:simsim} we plot the performance of $\VNA$ over a number of $(\ell,h)$ trimming pairs (we note that for all correlation/regularized/contami\-nated/trimmed combinations, mean performance is significantly better than chance and chance normalized plots are omitted).
In the left panel, we plot the modularity of the GMM clustering in the trimmed $G_2^{(\ell,h)}$ as a function of $\ell,h\in\{0,0.05,0.1,0.15,0.2,0.25\}$.
Note that we average the modularity values over $nMC=50$ randomly selected seed sets of size $s=10$ .
The color indicates the value of the modularity, with darker red indicating lower values and lighter yellow--white indicating larger values.
We see that modularity is maximized near $(\ell,h)\approx(0.1,0)$, and that the model-true trimming values $(\ell,h)=(0.1,0.1)$ achieves relatively high modularity as well.

\begin{figure*}[t!]
	\centering
	\subfloat[Regularized $\VNA$; $\rho=0.5$]  {\includegraphics[width=0.5\textwidth]{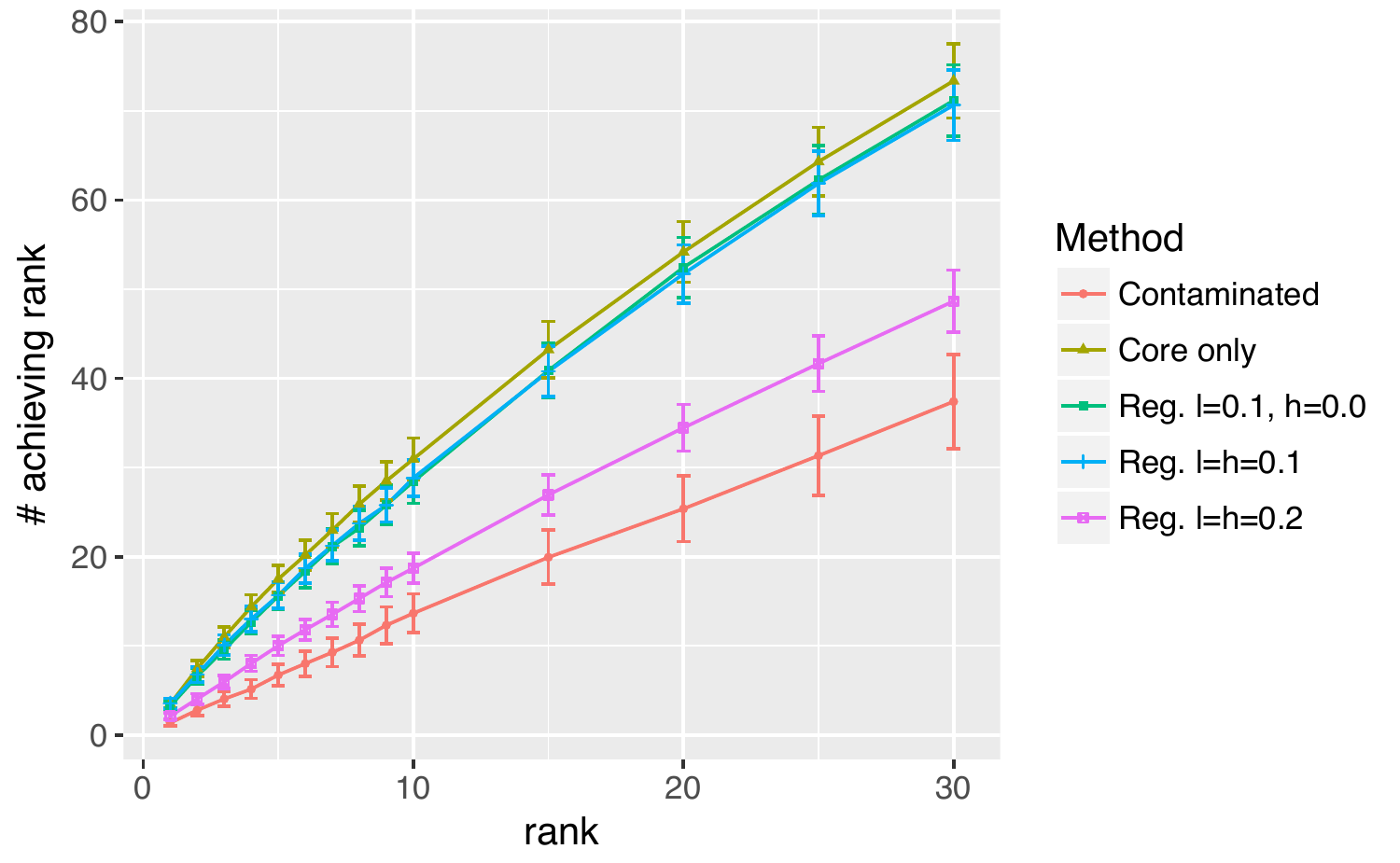}}
	\subfloat[Regularized $\VNA$; $\rho=0.3$]{\includegraphics[width=0.5\textwidth]{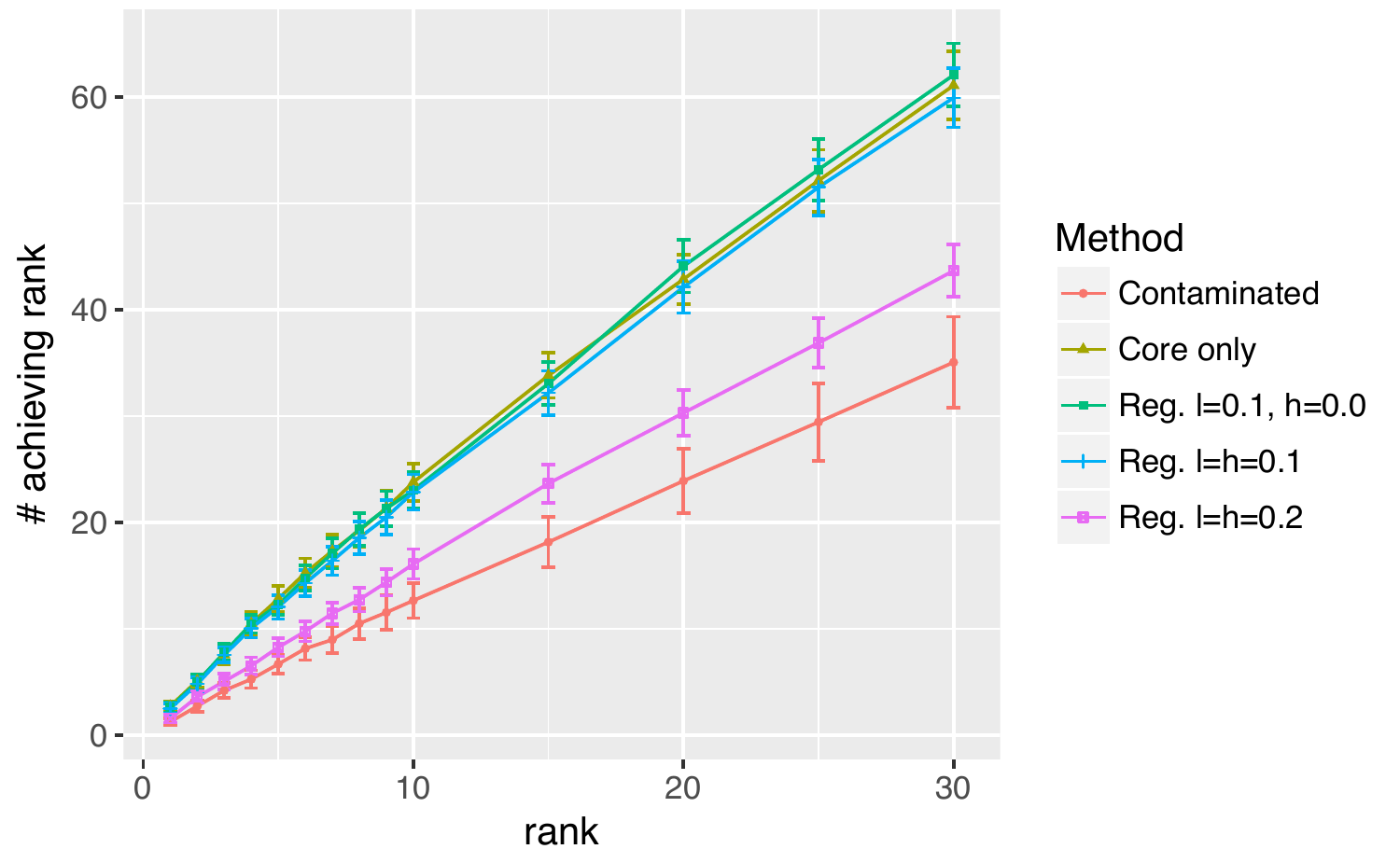}}

	\caption{
	In the right panel (resp., left panel), we plot
the performance of $\VNA$ in our contamination model in Section \ref{sec:model} with correlation $\rho = .5$ and $\rho = .3$ respectively, again averaged over $nMC=50$ random seed sets of size $s=10$.
The gold line represents the idealized network pair, the red represents the contaminated network pair, and the other colors represent various levels of regularization.
See Section \ref{sec: synth} for further details.
}
	\label{fig:simrho}
\end{figure*}

In the right panel, we plot
the performance of $\VNA$ ($\pm$2s.e.) in $(G_1,G_2)\sim\text{SBM}(0.7,200,2,B,\pi=(1/2,1/2))$ again averaged over $nMC=50$ random seed sets of size $s=10$.
The $x$-axis shows the ranks in the nomination list and the $y$-axis shows (on average) how many vertices $v\in G_1^{(i)}$, when viewed as the v.o.i., had their corresponding vertex of interest ranked in the top $x$ by $\VNA$. 
The gold line represents performance in the idealized network pair $(G_1^{(i)},G_2^{(i)})$; the red line for $(G_1^{(i)},G_2^{(c)})$; 
the green line for $(G_1^{(i)},G_2^{(0.1,0)})$;
the blue line for $(G_1^{(i)},G_2^{(0.1,0.1)})$; and
the pink line for $(G_1^{(i)},G_2^{(0.2,0.2)})$.
We see here that, as expected, performance loss due to contamination is mitigated by using the true model-based trimming parameters $\ell=h=0.1$, and using the modularity maximizing $\ell=0.1, h=0$.
If we over-trim, here represented by $\ell=h=0.2$, we see a degradation in performance; as expected from the low modularity value in the left panel for $\ell=h=0.2$.
We again see here the interesting phenomena observed in the motivating high school friendship network example of Section \ref{sec:Motive}:  modularity and subsequently VN performance tends to emphasize more trimming of the low degree vertices and less trimming of the high degree vertices.
This suggests that low-degree contamination is most effective at thwarting the performance on $\VNA$, perhaps contrary to the intuition that high-degree nodes adversely affect concentration of adjacency matrices \cite{le2017concentration}.

As in our motivating example, trimming can have the effect of removing v.o.i. from $G_2^{(c)}$, and we see this play out in Figure \ref{fig:simtrim}, in which we plot
the performance of $\VNA$ ($\pm$2s.e.) in $(G_1,G_2)\sim\text{SBM}(0.3,200,2,B,\pi=(1/2,1/2))$ with $\rho=0.7$ again averaged over $nMC=50$ random seed sets of size $s=10$.
The $x$-axis shows the ranks in the nomination list and the $y$-axis shows (on average) how many vertices $v\in G_1^{(i)}$, when viewed as the v.o.i., had their corresponding vertex of interest ranked in the top $x$ by $\VNA$. 
The gold line represents performance in the idealized network pair $(G_1^{(i)},G_2^{(i)})$; the red line for $(G_1^{(i)},G_2^{(c)})$; 
the green line for $(G_1^{(i)},G_2^{(0.1,0)})$;
the blue line for $(G_1^{(i)},G_2^{(0.1,0.1)})$; and
the pink line for $(G_1^{(i)},G_2^{(0.2,0.2)})$.

As expected, over-regularizing results in a significant number of v.o.i. being trimmed and significant performance loss as compared to the more moderate choices of regularization.
Lastly, exploring the affect of $\rho$ on $\VNA$ performance, we repeat the above experiment with $\rho=0.5,$ and $\rho=0.3$.
Results are plotted in Figure \ref{fig:simrho}.
As expected, the trends observed in Figure \ref{fig:simsim} hold here as well, with an across the board performance decrease as $\rho$ decreases.

\begin{figure*}[!t]
	\centering
\subfloat[ $\#$ achieving rank $\leq x$ versus $x$]{\includegraphics[width=0.5\textwidth]{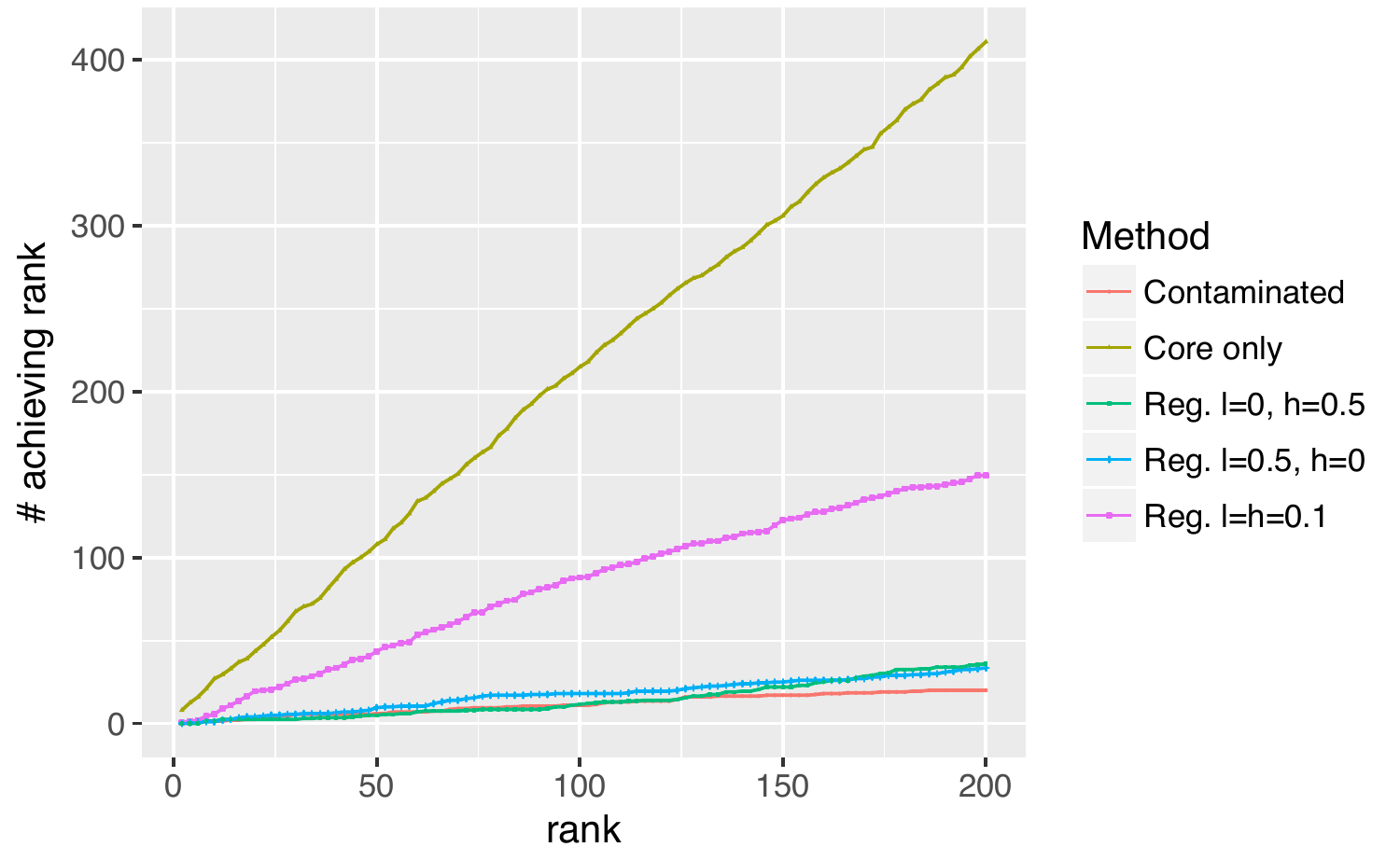}}
\subfloat[ Chance normalized $\#$ achieving rank $\leq x$ versus $x$]{\includegraphics[width=0.39\textwidth]{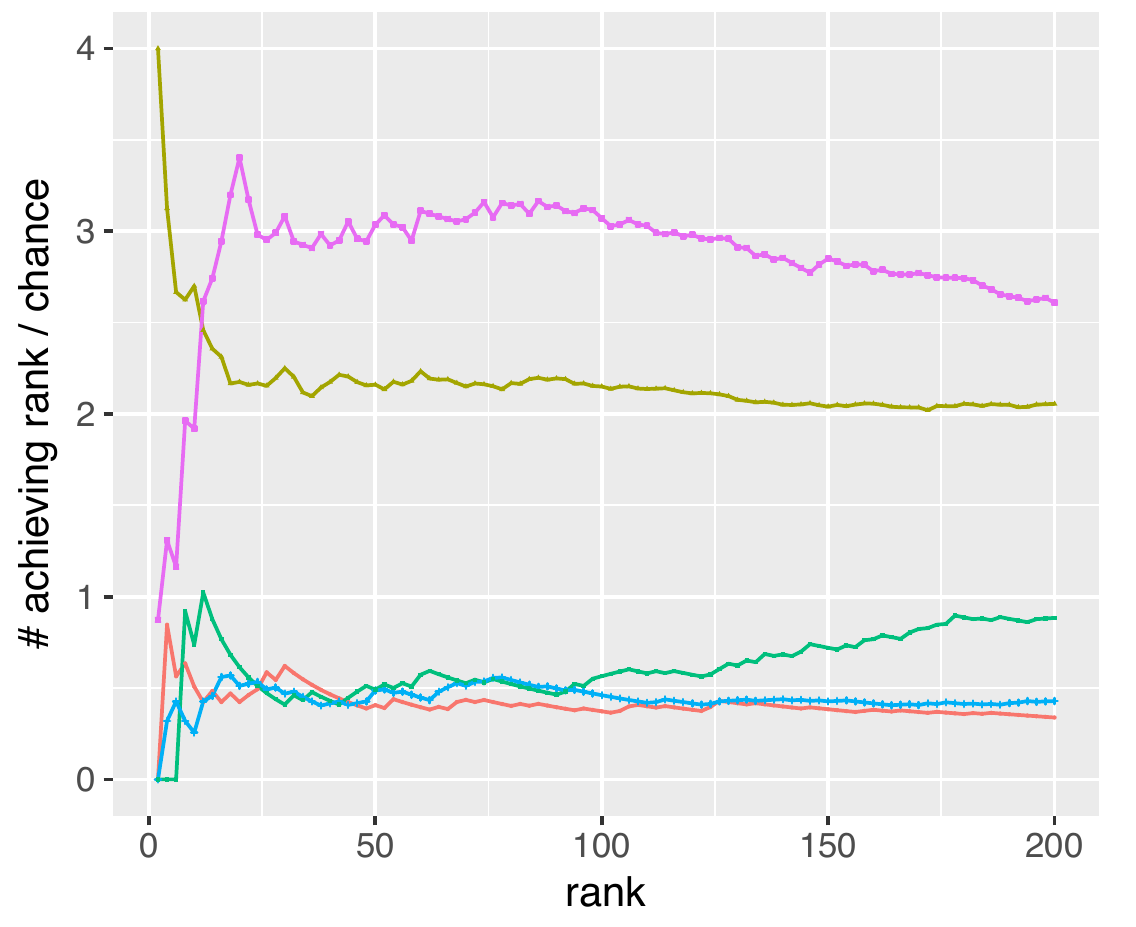}}
	\caption{
In panel a), we consider each vertex in $G_1^{(i)}$ as the v.o.i., and we plot the
number of vertices amongst these v.o.i. (x-axis) that had their corresponding
v.o.i. in $G_2$ ranked in the top $x$. 
The right panel shows the same result normalized by chance. We use 2 Monte Carlo replicates of $s = 100$ randomly chosen seeds, with the gold line representing the idealized network pair, the red line representing the contaminated, and the other colors representing various levels of regularization.  See Section \ref{sec:swim} for details.}
	\label{fig:bing}
\end{figure*}
\subsection{Microsoft Bing Entity Graph Transitions}
\label{sec:swim}

\begin{figure*}[t]
	\centering
	\subfloat{\includegraphics[width=0.6\textwidth]{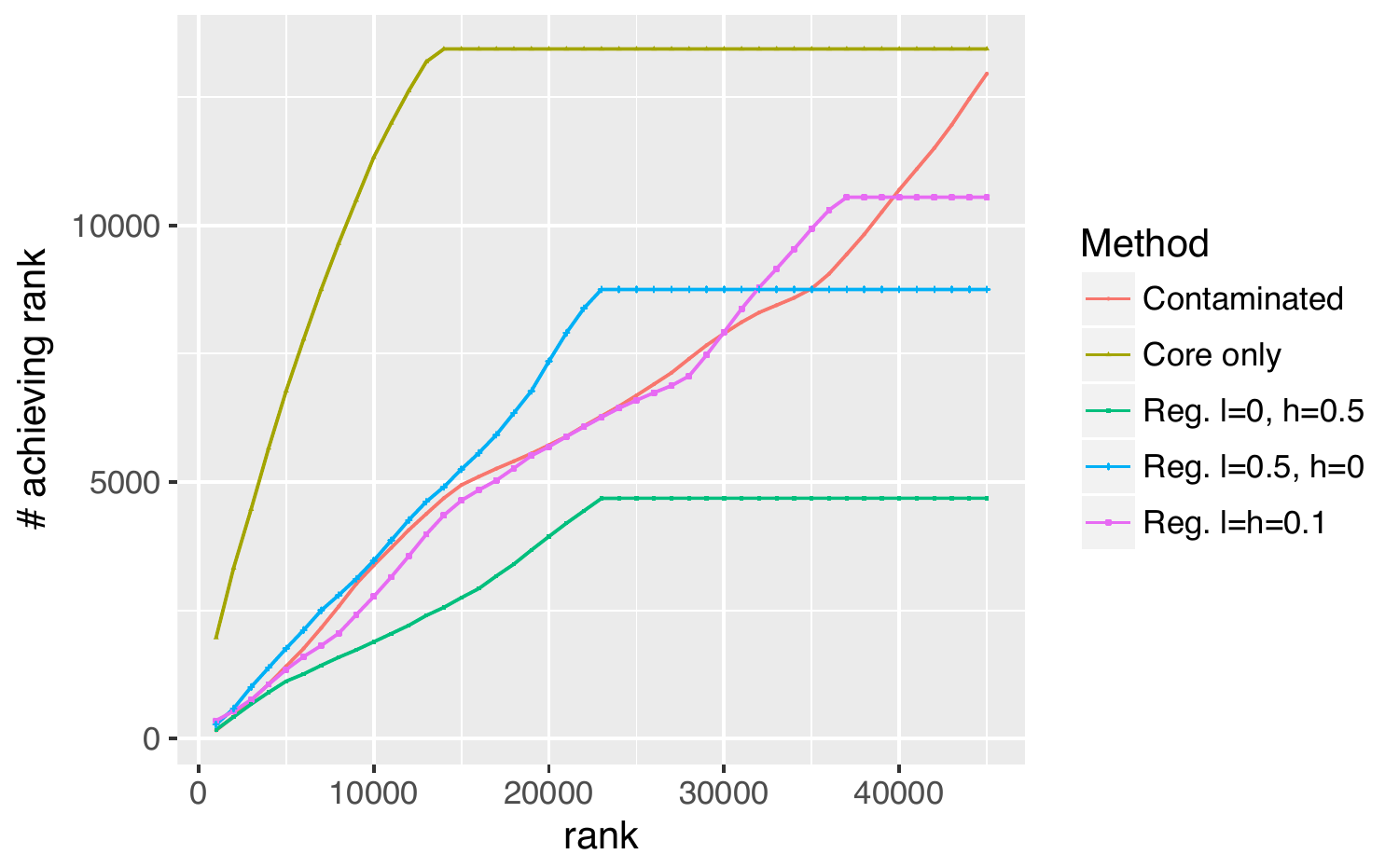} }
	\caption{
We consider each vertex in $G_1^{(i)}$ as the v.o.i., and we plot the
number of vertices amongst these v.o.i. (x-axis) that had their corresponding
v.o.i. in $G_2$ ranked in the top $x$. 
The right panel shows the same result normalized by chance. We use 2 Monte Carlo replicates of $s = 100$ randomly chosen seeds (with the same seed sets as in Figure \ref{fig:bing}), with the gold line representing the idealized network pair, the red line representing the contaminated, and the other colors representing various levels of regularization.  See Section \ref{sec:data} for details
}
	\label{fig:MSreg}
\end{figure*}

In the next example, we consider a multigraph derived from one month of aggregate Bing entity graph transitions.
The multigraph represents entity transitions, and  
each weighted edge-type of the multigraph represents aggregated signal that capture a transition rate between two entities while browsing.  
There are multiple ways that a transition between those entities could be made, so we count each aggregated signal separately using the different edge-types in the multigraph:  one edge-type represents transitions that were made via a suggestion interface;   
the other edge-type represents transitions that we made independent of any suggestion interface.  
As such, one type will have a constrained set of transition probabilities (it can realistically only connect to a subset of the vertices in the graph), while the other will be more ``unlimited'' in that it may connect to any other entity in the entire graph.

The resulting graphs are symmetric, weighted and loop-free, with $G_1^{(i)}$ containing $13535$ vertices and $519389$ edges,
$G_2^{(i)}$ containing $13535$ vertices and $595047$ edges, and the contaminated network
$G_2^{(c)}$ containing $45816$ vertices and $2848466$ edges.
Here, there is a 1-to-1 correspondence between the vertex sets of $G_1^{(i)}$ and $G_2^{(i)}$ with the contaminated network adding $32281$ vertices to $G_2^{(c)}$ that do not have a corresponding vertex in $G_1^{(i)}$.
In Figure \ref{fig:bing}, we explore the effect of this contamination (and the subsequent regularization) on $\VNA$.

Considering two randomly chosen sets of $s=100$ seeds, we run $\VNA$ on $(G_1^{(i)},G_2^{(i)})$ (yellow line in Figure \ref{fig:bing}), on $(G_1^{(i)},G_2^{(c)})$ (red line), on $(G_1^{(i)},G_2^{(0.1,0.1)})$ (pink line); on $(G_1^{(i)},G_2^{(0,0.5)})$ (green line); and on
	$(G_1^{(i)},G_2^{(0.5,0)})$ (blue line).
As in the simulations and motivating data example, we see the general trend of contamination adversely affecting performance and regularization ameliorating the effect of the contamination.
Here, the regularized graph $G_2^{(0.1,0.1)}$ has $36808$ vertices, and as expected, absolute performance (the left panel in Figure \ref{fig:bing}) in the clean case is better than in the regularized setting.
From the right panel, we observe however, that the relative improvement over chance achieved in the regularized setting exceeds that in the clean setting, and we observe that $\VNA$ performance is worse than chance in the contaminated and over-regularized network settings.
While regularization has not recovered the performance in the idealized setting, the improvement induced via regularization is dramatic versus the contaminated setting.
We also note that the modularity levels for automating the choice of $(\ell,h)$ in this example are relatively stable to the trimming value, with the clustered  $G_2^{(c)}$ achieving  $Q= 0.52$,
the clustered  $G_2^{(c)}$ achieving  $Q= 0.52$,
the clustered  $G_2^{(0,0.5)}$ achieving  $Q= 0.57$,
the clustered  $G_2^{(0.5,0)}$ achieving  $Q= 0.52$,
and 
the clustered  $G_2^{(0.1,0.1)}$ achieving  $Q= 0.53$.
Indeed, in this data example the graphs do not cluster particularly well under any trimming conditions, and a more modest trimming scheme is more effective for the  subsequent VN inference task.

In Figure \ref{fig:MSreg}, we again consider the performance of $\VNA$ with the same $nMC=2$ randomly chose 100 vertex seed sets and various levels of regularization, here plotting over an extended $x$-axis.
In pink we plot $\VNA$ run on $(G_1^{(i)},G_2^{(0.1,0.1)})$; in blue on $(G_1^{(i)},G_2^{(0.5,0)})$; in green on $(G_1^{(i)},G_2^{(0.3,0.3)})$; and in red on $(G_1^{(i)},G_2^{(0,0.5)})$.
This figure demonstrates another dramatic
side effect of over-regularization: v.o.i. that are trimmed for $G_2^{(c)}$ can never be recovered by $\VNA$.
This is represented by the horizontal asymptotes in Figure \ref{fig:MSreg}.

\section{Discussion}
\label{sec:discussion}
Our motivating question is two-fold:  What effect does adversarial contamination have on the performance of vertex nomination? 
Herein, we have demonstrated both theoretically and empirically that an adversary can cause our VN scheme to fail (i.e., nominate the wrong vertices).  
Empirically, we have also demonstrated that regularization can be effective for mitigating the effect of the contamination model posited herein, though we have not proven this result.
Establishing the theoretical effect of regularization on VN is an open problem, and the subject of our present research.


In \cite{lyzinski2017consistent}, the authors showed that there can be no universally consistent vertex nomination scheme assuming only one vertex of interest.  
In this paper, we have seen that with a suitable definition of a maximal consistency class and (possibly) multiple vertices of interest, there are infinitely many such consistency classes, which implies that ensemble methods cannot recover consistency and/or thwart an arbitrary adversary.   
This allows us to formulate our model of adversarial contamination in terms of consistency classes; indeed, an adversary for a particular VN rule aims to move the distribution out of the rule's consistency class.
A natural next question to consider would be what effect regularization has on a VN rule's consistency class.
Ideally, regularization enlarges the consistency class of a VN rule thereby making the adversary's job (i.e., moving the model out of the consistency class) more difficult.
The interplay between the adversary and regularization in VN is central to this story, although we are only at the infancy of understanding it.

\vv{There are several issues compounding the theoretical analysis of  regularization, even in the relatively simple setting posited herein.  
Indeed, the adversarially modified graph $G_2$ is, under our modeling assumptions of Section \ref{sec:model}, a stochastic blockmodel, albeit with more blocks than in $G_1$.
Theoretically analyzing the effect of our trimming regularizer in the context of $\VNA$ would require novel results in the concentration and spectral properties of regularized random graphs, akin (though different from) those in \cite{le2017concentration}.
Indeed, 
}
\textcolor{black}{regularization and its effect on the spectral analysis of random graphs is still not very well understood, as regularization often induces complicated dependency structure into the resulting regularized graph.  
Existing spectral analysis techniques often require relating differences in eigenvectors/eigenvalues} \vv{ for perturbed matrices with independent (or weakly dependent \cite{cape2019two}) entries, which is not directly applicable in the regularized setting.} \textcolor{black}{ Hence, new techniques must be developed to understand regularization.  We believe that our theoretical findings are a necessary first step to begin to understand how an adversary can affect vertex nomination.  
}
  
Our proposed definition of an adversary is suited to a general random graph setting, and it provides a simple surrogate in which to study the effect of contamination in real data examples.   
From our simulation study and real data examples we have seen that a particular VN rule ($\VNA$) succeeds before adversarial contamination, fails after contamination, and succeeds after graph regularization.  
We are currently exploring the effect of contamination on a broader class of VN rules, and 
considering other models for adversarial contamination and subsequent regularization. 
Finally, while we have partially answered in the negative our question about whether consistency can be retained in the general adversarial setting, another valid consideration is whether there are adversarial models for which the adversary does \textit{not} affect consistency.  
While we believe even simple manipulation on the edges of $G_2$ can affect consistency, it may be possible to derive bounds and phase transitions on the number of edges (or vertices) that an adversary would need to modify to change the result.  Mathematically, this is akin to finding limits on the size of $|V_{\mathcal{A}}|$ in our definition of an adversary.

\section*{Acknowledgements} 
This material is based on research sponsored by the Air Force Research Laboratory and DARPA under agreement number FA8750-18-2-0035.  This work is also supported in part by the D3M program of the Defense Advanced Research Projects Agency.  The U.S. Government is authorized to reproduce and distribute reprints for Governmental purposes notwithstanding any copyright notation thereon. The views and conclusions contained herein are those of the authors and should not be interpreted as necessarily representing the official policies or endorsements, either expressed or implied, of the Air Force Research Laboratory and DARPA or the U.S. Government.

\appendix
\section{Proof of Bayes Optimality for the Scheme in Sec. \ref{sec:BOcon}}
\label{sec:BO}

For each $i\in[h]$, $j\in[m]$, $v\in V^*$, $\Phi\in\mathcal{V}_{nm}$, define
\begin{align*}
U_{i,\mathbf{g}}^{j,v} :&= \bigg\{(g_1,g_2)\in \left(g_1^{(i)},[\mo(g_2^{(i)})]\right)\text{ s.t. } \rank_{\Phi(g_1,\mo(g_2),V^*)}(\mo(v)) =j\bigg\}\\
&=\bigg\{(g_1,g_2)\in \left(g_1^{(i)},[\mo(g_2^{(i)})]\right) \text{ s.t. }\Phi(g_1,\mo(g_2),V^*)[j]=\mo(v) \bigg\}\\
&=\Big\{(g_1,g_2)\in \left(g_1^{(i)},[\mo(g_2^{(i)})]\right) \text{ s.t. }\exists\text{ iso. }\sigma\text{ s.t. }\sigma(\mo(g_2^{(i)}))=\mo(g_2)\\
&\hspace{25mm} \text{ and } \sigma\left(\Phi(g_1^{(i)},\mo(g_2^{(i)}),V^*)[j]\right)= \mo(v)\Big\}\\
&=\left(g_1^{(i)},[\mo(g_2^{(i)})]\right)_{\Phi(g_1^{(i)},\mo(g_2^{(i)}),V^*)[j]=\mo(v)}.
\end{align*}
Lastly, for $(g_1,g_2)\in\gn^a\times \gm^a$, define $p_{\Phi} \in [0,1]^m$ via 
\begin{align*}
p^{(i)}_{\Phi}[g_1, \mo(g_2),V^*][j] &= p^{(i)}_{\Phi}[j] \\ :&= \sum_{v \in V^*} \PF \left[ U_{i,\mathbf{g}}^{j,v}\,\, \big|\,\, (g_1^{(i)},[\mo(g_2^{(i)})]
\right]\\
&= 
\PF \left[ E \right],
\end{align*}
where $E$ is the (conditional) event
\begin{align*}
    \bigg\{ (g_1,[\mo(g_2)])_{\Phi(g_1,\mo(g_2),V^*)[j] \in \mo(V^*)} \big| (g_1^{(i)},[\mo(g_2^{(i)}) \bigg\}
\end{align*}
Note that, by definition, $p_{\Phi^*}$ majorizes $p_{\Phi}$.

To show that $\Phi^*$ is Bayes optimal for $L^{(1)}_k$ (the proof for $L^{(2)}_k$ being completely analogous), we have that for $k\leq m-1$,
\begin{align*}
L^{(1)}_k(\Phi,V^*) &= 1-\frac{1}{|V^*|}\sum_{v\in V^*}\PF( \rank_{\Phi(G_1,\mo(G_2),V^*)}(\mo(v)) \leq k)\\
&=1-\frac{1}{|V^*|}\sum_{v\in V^*}\sum_{j\leq k}\PF( \rank_{\Phi(G_1,\mo(G_2),V^*)}(\mo(v)) = j)\\
&=1-\frac{1}{|V^*|} \sum_{\mathcal{P}_{\mathbf{g}}}\sum_{j\leq k}\sum_{v\in V^*}\bigg( \PF\left[ U_{i,\mathbf{g}}^{j,v}\,\big|\,  \left(g_1^{(i)},[\mo(g_2^{(i)})]\right)\right] \\ &\hspace{45mm} \times \PF\left[\left(g_1^{(i)},[\mo(g_2^{(i)})]\right)\right]\bigg) \\
&=1-\frac{1}{|V^*|} \sum_{\mathcal{P}_{\mathbf{g}}}\sum_{j\leq k}p^{(i)}_\Phi[j]\,\PF\left[\left(g_1^{(i)},[\mo(g_2^{(i)})]\right)\right]\\
&\geq 1-\frac{1}{|V^*|} \sum_{\mathcal{P}_{\mathbf{g}}}\sum_{j\leq k}p^{(i)}_{\Phi^*}[j]\,\PF\left[\left(g_1^{(i)},[\mo(g_2^{(i)})]\right)\right]\\
&=L^{(1)}_k(\Phi^*,V^*),
\end{align*}
as desired.

\section{Proof of Theorem \ref{lem:CC}}
\label{app:CC}
We first note that the growth condition on $|V_n^*|$ and on $k_n$ in the precision case ensures that the result for precision and recall consistency follow from each other, and so we will focus our attention on recall consistency.
The analogous result for precision follows mutatis mutandis.

Consider the following network construction for a network of size $n$.  
Let $\xi_n=\max(k_n,|V^*_n|)$.
For a fixed $p\in(0,1)$, let $B_1,\cdots,B_{\fl}$ be i.i.d. ER($\xi_n,p$) random graphs.
Let $H_{n}$ be a complete graph on $n-\xi_n\fl$ vertices.
Label the vertices  
\begin{align*}
&\text{ of }B_i\text{ with }\{1,2,3,\cdots,\xi_n\};\\
&\text{ of }B_2\text{ with }\{\xi_n+1,\xi_n+2,\xi_n+3,\cdots,2\xi_n\};\\
&\hspace{35mm}\vdots\\
&\text{ of }B_{i-1}\text{ with }\{(i-2)\xi_n+1,(i-2)\xi_n+2,\\&\hspace{40mm}(i-2)\xi_n+3,\cdots,(i-1)\xi_n\};\\
&\text{ of }B_1\text{ with }\{(i-1)\xi_n+1,(i-1)\xi_n+2,\\
&\hspace{40mm}(i-1)\xi_n+3,\cdots,i\xi_n\};\\
&\text{ of }B_{i+1}\text{ with }\{i\xi_n+1,i\xi_n+2,i\xi_n+3,\cdots,(i+1)\xi_n\};\\
&\hspace{35mm}\vdots\\
&\text{ of }B_{\fl}\text{ with }\bigg\{\left(\fl-1\right)\xi_n+1,\\&\hspace{20mm}\left(\fl-1\right)\xi_n+2,\cdots,\fl \xi_n\bigg\};\\
&\text{ of }H_n\text{ with }\left\{\fl \xi_n+1,\fl \xi_n+2,\ldots,n \right\}.
\end{align*}
For each $\ell\in\left[\fl\right]$ and each vertex $v$ in $V(B_\ell)$, independent of all other edges in the network, select $\ell$ vertices uniformly at random from $H_n$, i.e., from $$\left\{\fl \xi_n+1,\fl \xi_n+2,\ldots,n \right\}.$$
Denote this set of $\ell$ vertices via $V_{v,\ell}$---and place an edge between $v$ and each vertex in $V_{v,\ell}$.
Let $\mathcal{H}_{n,i}$ be the collection of all graphs possible under the above construction, and let $F_{n,i}$ be the distribution on $\mathcal{H}_{n,i}$ outlined above.

With $c=n$, the correspondence the identity, and (where $|V^*_n|=\nu_n$) $V^*_n = \{v_i\}_{i=1}^{\nu_n} = \{u_i\}_{i=1}^{\nu_n}=[\nu_n]$, define the collection of nominatable distributions 
$$\left\{\widetilde F_{n,i}\right\}_{i=1}^{\fl}$$ via $\widetilde F_{n,i}=F_{n,1}\times F_{n,i}$ (where ``$\times$'' denotes the usual product measure).

Suppose a VN rule $\boldsymbol\Phi=(\Phi_n)_{n=n_0}^\infty$ is level-$(k_n)$ recall consistent for $\mathbf{F}_i=(\widetilde F_{n,i})_{n=n_0}^\infty$.  
Then, by definition
$$\lim_{n \to \infty} L^{(1)}_{k_n} (\Phi_n, V^*) - L^{*,(1)}_{k_n}(V^*,\widetilde F_{n,i}) = 0.$$ 
However, note that here
$$L^{*,(1)}_{k_n}(V^*,\widetilde F_{n,i})\leq 1-\frac{k_n}{\xi_n}.$$  Indeed, for a given $\widetilde F_{n,i}$, consider the following VN scheme $\Psi_n$.
First identify the vertices of $H_n$; this is possible as $H_n$ is a complete subgraph of order $\geq 2n/3$, and each $B_i$ is of order $o(n)$ with vertices of degree at most $\fl\leq n/3$.
Each $B_\ell$ can then be recovered and identified by computing the number of edges between $H_n$ and each vertex $v\in V\setminus V(H_n)$; in particular $B_i$ can be identified as the set of vertices in $V\setminus V(H_n)$ with $i$ edges to $V(H_n)$.  Let $\psi_n$ then rank the vertices in $B_i$ (in arbitrary order) at the top of its nomination list.
It is immediate then that 
$$L^{(1)}_{k_n} (\Psi_n, V^*)=1-\frac{k_n}{\xi_n}.$$

\noindent By the distributional symmetry of the v.o.i., we have that for $v\in V^*$,
$$ 
\p_{\widetilde F_{n,i}}(\text{rank}_{\Phi_n(G_1, \fo (G_2), V^*)}(\fo(v)) \geq k_n+1)=L^{(1)}_{k_n}(\Phi_n,V^*).
$$
For any $\epsilon>0$ and sufficiently large $n$, consistency ensures that
$$\p_{\widetilde F_{n,i}}(\text{rank}_{\Phi_n(G_1, \fo (G_2), V^*)}(\fo(v)) \geq k_n+1) \leq \epsilon +\left(1-\frac{k_n}{\xi_n}\right).$$  
The internal consistency criterion in the definition of VN schemes (Eq. ref{eq:consis}), then implies that
\begin{equation}
\label{eq:contradicting_i}
\p_{\widetilde F_{n,i}}(\text{rank}_{\Phi_n(G_1, \fo (G_2), V^*)}(\fo(v)) \geq k_n+1) \leq \epsilon+\left(1-\frac{k_n}{\xi_n}\right)
\end{equation}
for each $v\in\{1,2,\cdots,\xi_n\}$.
Now, suppose that $\Phi$ is also level-$k_n$ recall consistent for ${\bf F}_j$ for $j \neq i$.  By similar logic, we must have that 
\begin{equation}
\label{eq:contradicting_j}
\p_{\widetilde F_{n,j}}(\text{rank}_{\Phi_n(G_1, \fo (G_2), V^*)}(\fo(v)) \geq k_n+1) \leq \epsilon+\left(1-\frac{k_n}{\xi_n}\right)
\end{equation}
for each $v\in\{1,2,\cdots,\xi_n\}$ for sufficiently large $n$.

Let $\sigma_{i\leftrightarrow j}$ be the permutation on $\{1, ..,n\}$ defined as 
\begin{align*}
\sigma(\ell ) := 
\begin{cases} 
(i-1)\xi_n + \ell & \ell \in \{ 1, 2, ...,\xi_n\} \\
\ell - (j-1)\xi_n & \ell \in \{(j-1)\xi_n + 1, ..., j\xi_n\} \\
(j-i)\xi_n + \ell & \ell \in \{(i-1)\xi_n + 1, ..., i\xi_n\} \\
\ell & \text{otherwise.}
\end{cases}
\end{align*}
Now, for each $v\in [\xi_n]$, define the sets
\begin{align*}
E_{n,i}^v :&= \{ (g_1,g_2) \in\mathcal{H}_{n,1}\times\mathcal{H}_{n,i}: \text{rank}_{\Phi_n(g_1, \fo (g_2), V^*_n)}(\fo(v)) \leq k_n \} \\
B_{n,i,j}^v :&= \{ (g_1,g_2) \in\mathcal{H}_{n,1}\times\mathcal{H}_{n,i}:  \text{rank}_{\Phi_n(g_1, \fo (g_2), V^*_n)}(\fo[(j-1)\xi_n+v]) \leq k_n \} \\
E_{n,j}^v :&= \{ (g_1,g_2) \in\mathcal{H}_{n,1}\times\mathcal{H}_{n,j}:  \text{rank}_{\Phi_n(g_1, \fo (g_2), V^*_n)}(\fo(v) \leq k_n \}
 \\ 
 E_{n,j} :&= \{ (g_1, \sigma(g_2)) \in \mathcal{G}_n\times\mathcal{G}_n : (g_1,g_2) \in E_{n,i} \} \\
 &= \{(g_1, \sigma(g_2)) \in \mathcal{G}_{n,j} : \text{rank}_{\Phi(G_1, o (G_2), v_1)}(o(\sigma(u_1))) \leq k \}. 
\end{align*}   
By consistency with respect to $\tilde F_{n,i}$ and $\tilde F_{n,j}$, i.e., by Eqs. \ref{eq:contradicting_i}--\ref{eq:contradicting_j}, we have that for any $\epsilon>0$, there exists $\tilde n$ such that for $n\geq \tilde n$, we have 
\begin{align}
\label{eq:goodE}
\p_{\widetilde F_{n,i}}( E^v_{n,i}) &\geq \frac{k_n}{\xi_n}-\epsilon;  \\
\p_{\widetilde F_{n,j}}(E^v_{n,j})&\geq \frac{k_n}{\xi_n}-\epsilon.\notag
\end{align} 
As $(G_1,G_2)\sim \widetilde F_{n,i}\Leftrightarrow(G_1,\sigma(G_2))\sim \widetilde F_{n,j}$, the 
\begin{align}
\label{eq:badB}
\p_{\widetilde F_{n,j}}(E^v_{n,j})=\p_{\widetilde F_{n,i}}( B^v_{n,i,j}) \geq \frac{k_n}{\xi_n}-\epsilon.
\end{align}
For each $v\in[\xi_n]$ and $h\in[k_n]$ and $i \in [n]$, define the sets
\begin{align*}
    R_{i,v,h} :&= \bigg\{ (g_1,g_2) \in\mathcal{H}_{n,1}\times\mathcal{H}_{n,i}:  \text{rank}_{\Phi_n(g_1, \fo (g_2), V^*_n)}(\fo(v)) =h \bigg\}\\
\end{align*}
Then, define
\begin{align*}
\alpha_{v,h}=&\p_{\widetilde F_{n,i}}\big[\,R_{i,v,h}\,\big];\\
\beta_{v,h}=&\p_{\widetilde F_{n,i}}\big[\,S_{i,v,h}\,\big].
\end{align*}
By Eq. \ref{eq:goodE}, we have that $\sum_{h=1}^{k_n}\alpha_{v,h}\geq \frac{k_n}{\xi_n}-\epsilon$, and
by Eq. \ref{eq:badB}, we have that $\sum_{h=1}^{k_n}\beta_{v,h}\geq \frac{k_n}{\xi_n}-\epsilon$.
Noting that for each $h\in[k_n]$
\begin{align*}
1&\geq \p_{\widetilde F_{n,i}}\left[\left(\cup_{v\in \xi_n]}R_{i,v,h}\right)\cup\left(\cup_{v\in [\xi_n]}S_{i,v,h} \right)\right]\\
&=\sum_{v\in[\xi_n]}\p_{\widetilde F_{n,i}}(R_{i,v,h})+\p_{\widetilde F_{n,i}}(S_{i,v,h})\\
&=\xi_n\alpha_{v,h}+\xi_n\beta_{v,h},
\end{align*}
and hence $$\beta_{v,h}\leq \frac{1}{\xi_n}-\alpha_{v,h}.$$
Plugging this into Eq. \ref{eq:badB} then yields
\begin{align*}
\frac{k_n}{\xi_n}-\epsilon&\leq=\p_{\widetilde F_{n,i}}( B^v_{n,i,j})\\
&=\sum_{h=1}^{k_n}\beta_{v,h}\\
&\leq \frac{k_n}{\xi_n}-\sum_{h=1}^{k_n}\alpha_{v,h}\\
&\leq \epsilon.
\end{align*}
As $\epsilon$ was chosen arbitrarily, and $\frac{k_n}{\xi_n}$ is bounded away from 0 by assumption, we reach our desired contradiction, and $\Phi$ cannot be consistent with respect to both ${\bf F}_i$ and ${\bf F}_j$.
As $i,j\in\lfloor \frac{n_0/3}{\xi_{n_0}}\rfloor$ were arbitrary, we see that there must be at least countably many consistency classes (since there are at least $\lfloor \frac{n_0/3}{\xi_{n_0}}\rfloor$ and we can let $n_0$ tend to infinity).

\bibliographystyle{plain} 
\bibliography{biblio.bib,adversary.bib} 
 \end{document}